\documentclass[journal]{IEEEtran}
\usepackage{amsmath,amsfonts}
\usepackage{algorithmic}
\usepackage{array}
\usepackage{textcomp}
\usepackage{stfloats}
\usepackage{url}
\usepackage{verbatim}
\usepackage{cite}               % order multiple entries in \cite{...}
\usepackage{xcolor}             % color definitions, to be use for...
\usepackage[]{hyperref}         % ...clickable refs within pdf...
\usepackage{tikz}
\usepackage{amsmath}
\usepackage{graphicx} %插入图片的宏包
\usepackage{float} %设置图片浮动位置的宏包
\usepackage{algorithm}
\usepackage{subfigure}
\usepackage{booktabs}
\usepackage[most]{tcolorbox}
\usepackage{multirow}
\usepackage{enumitem}
\usepackage{siunitx}

\def\BibTeX{{\rm B\kern-.05em{\sc i\kern-.025em b}\kern-.08em
    T\kern-.1667em\lower.7ex\hbox{E}\kern-.125emX}}
\usepackage{balance}

\begin{document}
\title{Malicious Image Analysis via Vision-Language Segmentation Fusion: Detection, Element, and Location in One-shot}

\author{
\IEEEauthorblockN{
Sheng Hang$^1$, Chaoxiang He$^1$, Hongsheng Hu$^1$, Hanqing Hu$^1$,\\
Bin Benjamin Zhu$^2$, Shi-Feng Sun$^1$, Dawu Gu$^1$\IEEEauthorrefmark{1}, Shuo Wang$^1$\IEEEauthorrefmark{1}
\thanks{\IEEEauthorrefmark{1}Corresponding authors.}
}

\IEEEauthorblockA{
$^1$Shanghai Jiao Tong University, Shanghai, China; \\
$^2$Microsoft Corporation, China\\
}
}

\markboth{Journal of \LaTeX\ Class Files,~Vol.~18, No.~9, September~2020}%
{How to Use the IEEEtran \LaTeX \ Templates}

\maketitle

\begin{abstract}
Detecting illicit visual content demands more than image-level NSFW flags; moderators must also know what objects make an image illegal and where those objects occur. We introduce a zero-shot pipeline that simultaneously (i) detects if an image contains harmful content, (ii) identifies each critical element involved, and (iii) localizes those elements with pixel-accurate masks—all in one pass. The system first applies foundation segmentation model (SAM) to generate candidate object masks and refines them into larger independent regions. Each region is scored for malicious relevance by a vision-language model using open-vocabulary prompts; these scores weight a fusion step that produces a consolidated malicious object map. An ensemble across multiple segmenters hardens the pipeline against adaptive attacks that target any single segmentation method.

Evaluated on a newly-annotated 790-image dataset spanning drug, sexual, violent and extremist content, our method attains  
\SI{85.8}{\percent} element-level recall, \SI{78.1}{\percent} precision and a \SI{92.1}{\percent} segment-success rate—exceeding direct zero-shot VLM localisation by \SI{27.4}{\percent} recall at comparable precision.  
Against PGD adversarial perturbations crafted to break SAM and VLM, our method's precision and recall decreased by no more than \SI{10}{\percent}, demonstrating high robustness against attacks.
The full pipeline processes an image in seconds, plugs seamlessly into existing VLM workflows, and constitutes the first practical tool for \emph{fine-grained, explainable} malicious-image moderation. 
\end{abstract}

\begin{IEEEkeywords}
Vision-Language Models (VLM), Harmful Content Localization, Visual Content Safety, Open-Vocabulary Detection.
\end{IEEEkeywords}

\textcolor{red}{\textbf{Warning}: The paper contains content that may be viscerally uncomfortable and not safe for work (NSFW).}

\section{Introduction}

\IEEEPARstart{T}HE detection of malicious or harmful visual content – such as depictions of drug use, explicit sexual activity, or violent acts – is a crucial problem for content moderation and security. Traditional NSFW image classifiers typically assign a coarse label (e.g., “safe” or “not safe”) or broad category to an entire image. While such classifiers (often based on CNNs or CLIP models) can identify obvious adult or graphic content~\cite{nian2016pornographic, tabone2021pornographic, radford2021clip}, they offer little insight into what specific objects or actions make the image unsafe. Moreover, many malicious scenarios are recognizable only through the combination of multiple innocuous objects. For instance, a photograph containing a human figure, a medicine bottle, and white powder might jointly imply an illicit drug use event – none of these objects alone is definitively illegal, but their co-occurrence is suspicious. Recognizing these fine-grained contextual cues exceeds the capabilities of standard NSFW classifiers and demands a more detailed visual understanding.

Fine-grained malicious element detection is a specialized image understanding task that identifies key harmful objects (e.g., weapons, paraphernalia) and their relationships. While related to Human-Object Interaction (HOI) detection and co-occurrence analysis~\cite{chao2018hoi,yatskar2016situation}, our approach is broader. Unlike HOI, we detect complex scenarios like "drug abuse" rather than simple, predefined interactions. Different from general co-occurrence analysis, we specifically focus on object combinations that signal malicious content, not just any frequent pattern.

% Fine-grained malicious element detection can be viewed as a specialized case of high-level image understanding that intersects with object detection, scene understanding, and human-object interaction recognition. In particular, it requires identifying key objects (e.g., paraphernalia, weapons, explicit body parts) and understanding their relations (e.g., a person holding a syringe). This is related to tasks like human-object interaction (HOI) detection, which aims to localize humans and objects and classify their interaction~\cite{chao2018hoi}. However, our goal extends beyond predefined interaction verbs – we seek to detect a broader notion of a scenario (such as “drug abuse” or “violent assault”) that may involve multiple objects and actors. It also relates to co-occurrence analysis in multi-object scenes, where patterns of frequently co-occurring objects are used to infer higher-level context~\cite{yatskar2016situation}. Unlike general co-occurrence mining, here the co-occurring elements are specifically those that signal malicious or prohibited content.

Recent advances in foundation models for vision offer new tools to tackle this challenge. On one hand, foundation segmentation models like SAM (Segment Anything Model) have demonstrated impressive zero-shot ability to partition an image into objects or regions without task-specific training\cite{kirillov2023segment}. SAM can generate high-quality object masks given minimal prompts, or even segment “everything” in an image, making it a powerful starting point for discovering all objects present.
% Extensions such as Grounded-SAM combine a zero-shot detector (Grounding DINO) with SAM to enable open-set, text-driven object segmentation\cite{liu2024groundingdino}. Another model, SEEM (Segment Everything Everywhere All at Once), provides a universal interface for segmentation, supporting various prompt types (points, boxes, text) in a single model\cite{zou2023seem}. 
These models can produce candidate object masks for any item in the scene, including potentially malicious elements, even if those objects were never explicitly seen in training – a significant advantage for a zero-shot detection setting.

On the other hand, large-scale vision-language models (VLMs) have dramatically improved in visual understanding capabilities. Models such as Qwen-VL\cite{bai2025qwen} and BLIP-2\cite{li2023blip2} combine vision encoders with language models, enabling them to not only caption images or answer questions, but also to localize and identify objects in an open-vocabulary manner. For example, Qwen-VL is designed to accept images and output text and even bounding boxes, supporting fine-grained recognition and localization of described entities. These VLMs leverage knowledge from multimodal training (image-text pairs) to recognize a vast array of visual concepts and can make zero-shot predictions about whether a given image (or region) contains instances of a concept (e.g., “drug paraphernalia” or “blood”) simply by being prompted in natural language. However, generic VLMs alone might overlook small yet crucial details if the prompt is not specific, and their outputs (like captions) might not highlight where in the image the malicious content is.

\begin{figure}[!ht] 
\centering 
\includegraphics[width=0.49\textwidth]{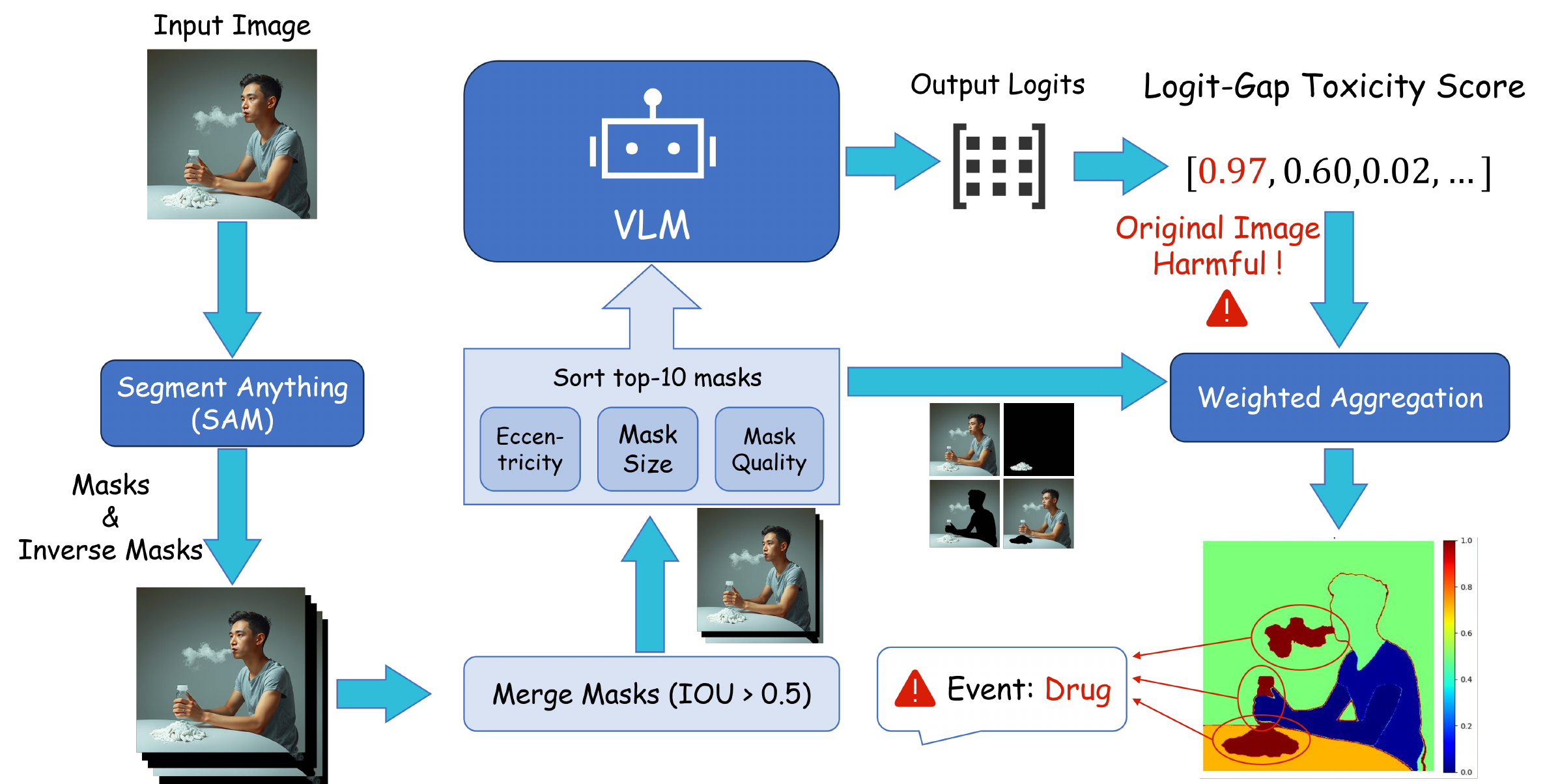}
\caption{Overall architecture of our pipeline} 
\label{fig:pipeline} 
\end{figure}

Considering the above, a promising strategy is to marry the strengths of segmentation models and VLMs: use segmentation to enumerate and isolate all candidate objects in the image, and use a VLM to semantically assess each object’s role in a malicious scenario. This forms the backbone of our proposed method. By scoring segmented objects with a VLM, we can identify which objects (or object sub-parts) are likely associated with a malicious event. Furthermore, by projecting those scores back onto the image (via the object masks), we obtain a spatial map of the malicious content – effectively “lighting up” the exact regions (e.g., a syringe, a weapon, drugs) that cause the image to be flagged. This fine-grained explanatory power is valuable for human moderators and for further automated processing (such as blurring or removing specific elements).

\noindent \textbf{Challenges.} Several challenges arise in this approach. First, segmentation models like SAM, while powerful, can over-segment or under-segment objects. A single mask may cover multiple adjoining objects or miss fine components. We address this with a sub-object mask refinement step to break down composite segments. Second, a purely zero-shot approach must contend with the open-ended variety of “malicious” content. We cannot predefine all malicious objects; instead, we rely on the VLM’s broad knowledge and use a logit-based scoring to indicate malicious likelihood without hard-coded categories. Third, adversarial or adaptive attackers could attempt to exploit the segmentation stage – for example, by camouflaging contraband so that segmentation models fail to isolate it, or by adding perturbations that cause masks to fragment. Recent studies have indeed found that even SAM is vulnerable to adversarial perturbations, which can remove or alter masks with imperceptible changes to the image\cite{zhang2023attacksam, zhou2024darksam}. This motivates our use of multiple segmentation models in tandem and a aggregation strategy to improve resilience: an attacker would have to fool all segmentation models simultaneously, which is significantly harder.

\noindent \textbf{Contributions.} We make the following contributions: 
\begin{itemize}[leftmargin=*]
    \item We present a novel pipeline for malicious image detection that identifies not just the presence of unsafe content but the specific objects and their locations contributing to it. 
    % To our knowledge, this is the first method to tackle fine-grained recognition of malicious elements via object co-occurrence patterns.

    \item We integrate foundation segmentation (SAM) with a vision-language scoring mechanism (using a model like Qwen-VL) to achieve open-vocabulary, zero-shot detection of arbitrary suspicious objects. The approach requires no task-specific training, leveraging pre-trained models for both segmentation and semantic scoring.

    \item We propose an integration strategy across segmentation models as a defense against segmentation-targeted attacks, and analyze its effectiveness against adaptive attackers in a zero-shot scenario. 

    \item We evaluate our method on a newly curated dataset of real NSFW images with element-level annotations (covering illegal activities, sexual content, extremism and violence). The results show clear improvements in segment success rate, element detection rate, and boundary alignment over a range of baselines: Grad-CAM explanations on classifier/VLM, zero-shot Qwen-VL segmentation outputs, attention maps on Qwen-VL. Our approach also serves as a plugin module to enhance existing VLMs with malicious content detection capability, combining their semantic understanding with a dedicated detection mechanism.
\end{itemize}
\section{Preliminary}
\subsection{Related Work}
\noindent \textbf{Harmful Content Classification.} Early work on detecting not-safe-for-work (NSFW) or illicit images focused on image-level classification using deep learning. Convolutional neural networks trained on adult-content datasets can achieve high accuracy in flagging pornography or gore\cite{nian2016pornographic, tabone2021pornographic}. For instance, Nian et al. employed a deep CNN to classify images into explicit vs. safe categories\cite{nian2016pornographic}, and subsequent systems improved on nudity detection using modern architectures and transfer learning\cite{tabone2021pornographic}. Transformer-based models and open-source safety classifiers (e.g., used in Stable Diffusion and LAION filters) have also been explored\cite{radford2021clip}. However, these classifiers output only a coarse label and do not indicate which part of an image is problematic. They struggle with context-dependent scenarios where no single object is obviously illicit. This limitation motivates approaches that go beyond whole-image classification to provide element-level identification and explanation.

%TODO
% \noindent \textbf{Object Detection for Unsafe Elements.} An intuitive extension is to detect specific dangerous or explicit objects within images. Researchers have developed object detectors for weapons, drugs, and other contraband. For example, Salido et al. trained a deep model to automatically detect handguns in surveillance images\cite{salido2021handgun}. Such object-specific detectors can localize items like firearms or knives in an image. Likewise, algorithms exist for detecting blood or explicit body parts using region proposals and CNNs. These methods improve interpretability by pinpointing where a risky object is, but typically each model is limited to a predefined category (e.g., “gun”). They lack the flexibility to handle arbitrary unseen malicious objects. Moreover, many illicit situations involve combinations of benign objects – a gap which per-object detectors cannot bridge. Our work addresses this by using an open-vocabulary detector (via a VLM) that can flag novel suspicious items and, importantly, reason over multiple objects jointly. 

\noindent \textbf{Context and Interaction Understanding.} Recognizing harmful content requires understanding contextual cues, similar to tasks like Human-Object Interaction (HOI) detection~\cite{chao2018hoi}, scene graph generation, and situation recognition~\cite{yatskar2016situation}, all of which emphasize that object relationships are key to understanding complex events. However, these prior methods typically rely on closed-set vocabularies of predefined actions or situations, making them ill-suited for the diverse nature of malicious content. In contrast, our approach leverages an open-world Vision-Language Model (VLM) to identify novel and unconventional object combinations that signal a harmful event, such as a "person + powder + spoon" arrangement indicating drug use.

% The importance of contextual cues in harmful content has parallels in tasks like visual relationship detection and HOI detection. HOI detectors localize pairs of humans and objects to recognize interactions (e.g., “person holding syringe”)\cite{chao2018hoi}, and scene graph models label relationships among objects (e.g., “knife on table”). Situation recognition goes further to label the overall activity or scenario depicted, along with roles of objects\cite{yatskar2016situation}. These research lines underscore that recognizing complex events (like a violent or illicit act) requires understanding how entities relate. However, prior HOI and situation-recognition methods were not designed specifically for malicious content and often rely on closed-set vocabularies of interactions or situations. In contrast, our approach leverages open-world knowledge in a VLM to identify unconventional or previously unseen combinations of objects that signal a malicious event (e.g., a person + powder + spoon arrangement indicating drug preparation). 

\noindent \textbf{Vision-Language Models (VLMs) for Moderation.} With the rise of foundation models, recent works have explored multimodal and zero-shot techniques for content moderation. The Hateful Memes Challenge\cite{Memes} introduced a benchmark for detecting harmful multimodal content, spurring models that combine vision and language understanding. For instance, Pro-Cap by Cao et al. uses a frozen VLM (CLIP-based) to identify hateful meme content\cite{cao2023procap}, and MOMENTA employs a multimodal reasoning framework for harmful memes\cite{pramanick2021momenta}. More directly related to our goals, Guo et al.\cite{guo2024moderating} propose using a VLM to detect illicit image promotions in online games, and Bao et al.’s VModA framework integrates a VLM with an LLM to moderate NSFW images in a zero-shot, adaptive manner\cite{bao2025vmoda}. VModA introduces clever prompting strategies (e.g., region zoom-in and chain-of-thought) to improve a VLM’s focus on image details. 

Similar to our insight, Reasoning Segmentation Models leverage both SAM and VLM to segment specific object according to given prompt, offering a powerful tool for nuanced understanding in image segmentation and recognition. For instance, LISA~\cite{lai2024lisa} and LLM-Seg~\cite{wang2024llmseg} inputs the specific tokens output by the VLM and the segmentation options provided by SAM into a trained fusion model to achieve the reasoning segmentation task. However, these works do not operate in a zero-shot manner and falls short when it comes to identifying multiple objects and elements, such as multiple nude people or multiple scattered bloodstains. In contrast, our work remains zero-shot, as it does not require any training. By explicitly generating segmentation masks for malicious elements and giving a heatmap of the harmfulness level with clear gradations, our method does not rely on language descriptions alone or giving result choosing one object. This yields visual explainability and allows fine-grained localization of the offending content. 
% To our knowledge, our system is the first to fuse segmentation with VLM scoring for content moderation, enabling it to surface which pixels correspond to unsafe content in an image.

\noindent \textbf{Adversarial Robustness and Explainability.} Security researchers have shown that content moderation models can be vulnerable to evasion. Adversarial attacks can subtly alter an image so that a classifier fails to detect nudity or violence. In the context of segmentation, recent studies Attack-SAM\cite{zhang2023attacksam} and DarkSAM\cite{zhou2024darksam} demonstrated that even the Segment Anything Model can be fooled into missing or falsely segmenting objects via crafted perturbations. These findings underline the need for resilient moderation techniques. Our ensemble of segment model and VLM is a direct response to this concern, making successful attacks less likely. Finally, explainable AI techniques are crucial for moderator trust. Traditional saliency methods like Grad-CAM\cite{selvaraju2017gradcam} have been applied to NSFW classifiers to highlight important image regions, but they produce coarse heatmaps and cannot distinguish specific objects. By providing pixel-accurate masks and labels for malicious elements, our approach offers a more interpretable output that moderators (or automated systems) can act upon—blurring, removing, or scrutinizing the highlighted regions as needed.

% no \IEEEPARstart
\subsection{Segmentation and VLMs}
Foundation Segmentation Models. Object segmentation has long been a fundamental task in computer vision. Earlier approaches required task-specific training for a fixed set of classes (e.g., Mask R-CNN\cite{maskrcnn} for COCO\cite{coco} categories). Recent foundation models like SAM (Segment Anything) have revolutionized this area by introducing generic, promptable segmentation trained on billions of masks. SAM can accept prompts (points, boxes, or text queries via a separate classifier) to segment target regions, or it can automatically produce a set of masks covering an image. It achieves impressive zero-shot transfer to new domains, often matching or exceeding task-specific models. However, SAM on its own does not assign semantic labels to its masks – it doesn’t “know” what each segment is. For identifying malicious content, we need to bridge this gap with semantic understanding.

% To inject semantic awareness into segmentation, researchers have proposed combining SAM with open-vocabulary detection models. Grounded-SAM is one such framework that uses Grounding DINO (an open-set object detector driven by text prompts) to first detect objects described by a text label, then applies SAM to obtain precise masks for those detections. Grounded-SAM essentially allows segmentation “of any object” given its name, enabling a form of zero-shot semantic segmentation. It achieved strong results on open-vocabulary segmentation benchmarks, e.g., a 48.7 mAP on the Segmentation-in-the-Wild (SegInW) dataset using Grounding DINO plus SAM. Another relevant model is SEEM (Segment Everything Everywhere All at Once), which takes a more unified approach: SEEM is a single model that accepts multiple types of prompts (including text) and produces the corresponding segmentation mask(s) in one forward pass. It’s designed as a “universal segmentation interface” analogous to how LLMs serve as universal text interfaces. SEEM’s versatility (handling points, boxes, scribbles, and text queries together) and its semantic-awareness via a text encoder make it particularly attractive for our problem, since it can directly respond to queries like “segment all medicine bottles in the image” or even more abstract prompts. In our pipeline, we will leverage SAM for its strong general segmentation and consider Grounded-SAM and SEEM as complementary segmentation front-ends when integrating multiple models.

Although existing tools like Grounded SAM provide more advanced semantic understanding, their highly specific approach to element extraction can be restrictive for abstract tasks like harmful content detection. This is particularly true for identifying amorphous or sensitive elements, such as irregular bloodstains and private body parts. Consequently, we utilize the native Segment Anything Model (SAM) and feed multiple masked segments into a VLM to facilitate a more comprehensive understanding.

VLMs for Image Understanding. The rise of large-scale VLMs has provided powerful tools that jointly understand images and text. Qwen-VL-7B is an example of a state-of-the-art VLM: it extends a 7-billion-parameter language model with visual encoding and was trained to perform a wide range of multimodal tasks, including captioning, visual question answering, and grounding. Notably, Qwen-VL can output not just text (e.g., answers or descriptions) but also bounding boxes, making it capable of referring expression comprehension – given a phrase like “the man holding a knife,” it can identify the corresponding region in the image. This ability is crucial for zero-shot localization of arbitrary objects. The Qwen-VL series (which includes an instruct-tuned chat model) has achieved record-breaking results for generalist models on benchmarks spanning image captioning, VQA, and visual grounding. Its multimodal knowledge base allows it to recognize a vast array of objects and even read text in images (through an integrated OCR mechanism).

% Another influential model, BLIP-2, introduced a strategy to connect frozen image encoders with large language models via a learned bridge. BLIP-2 demonstrated that high-performing image captioning and QA could be achieved by leveraging pre-trained components, achieving state-of-the-art results with much fewer trainable parameters than end-to-end training. We mention BLIP-2 as it represents the class of VLMs that can be adapted to new tasks efficiently. In our context, a model like BLIP-2 (or its instruction-tuned variant InstructBLIP) could describe an image and possibly highlight suspicious content in the description. However, descriptions alone are insufficient for pinpointing malicious elements – hence the need to combine such models with segmentation.

It is worth noting that simpler multimodal approaches like CLIP (Contrastive Language-Image Pretraining) have also been applied to NSFW detection: for example, by embedding an image and measuring similarity to text concepts like “porn” or “drugs,” one can flag content without explicit training. In fact, one NSFW filtering method computes CLIP embedding distances to a set of 17 NSFW concept words to decide if an image is unsafe. While effective for broad filtering, such methods still do not localize offending content. Our use of VLMs goes a step further: we effectively ask the model about each specific region, “Is this part of a [malicious event]?” and thus obtain both a semantic judgment and a spatial localization.

\subsection{Threat Model}
\noindent \textbf{Zero-Shot Setting}. We operate in a zero-shot paradigm: the models (segmenters and VLM) are pre-trained on general data and not fine-tuned for the malicious detection task. This is a realistic scenario for real-world deployment, since collecting and labeling a large dataset of malicious images with pixel-wise annotations is difficult and costly. However, this means our system must rely on the inherent capabilities of foundation models. The threat model assumes that potential attackers (e.g. users attempting to upload disallowed content or adversarially manipulate images) are aware of the detection system and may try to circumvent it. Two levels of adversaries are considered:

\noindent \textbf{Passive (Zero-Shot) Adversary}: This refers to malicious content that wasn’t specifically modified to evade our system, but may still be challenging due to distribution shift or complexity. For example, an image might depict a novel way of concealing drugs that the models haven’t seen, or an angle/lighting that makes segmentation hard. The system should handle these as well as possible with zero-shot knowledge.

\noindent \textbf{Adaptive Adversary}: This stronger adversary deliberately alters the image to attack the segmentation stage (since that is an obvious first point of failure). Inspired by recent research on SAM’s adversarial robustness, an attacker might add imperceptible perturbations to an image to cause the segmentation model to either miss an object or produce a misleading mask. For instance, an attacker could perturb pixels around a syringe so that SAM no longer generates a mask tightly around it (mask removal or fragmentation attack).

Likewise, the downstream VLM stage can also be targeted. Drawing inspiration from VLM jailbreaking, an attacker could craft an adversarial perturbation on a harmful image, thereby deceiving the VLM into classifying it as harmless.

Our threat model does not assume the attacker can directly tamper with the internal model (no model evasion beyond input manipulation), and we assume they do not have unlimited queries (to prevent, say, exhaustive search for adversarial inputs). However, we do consider that they might know which segmentation model or VLM is in use and adapt specifically to its vulnerabilities.

\noindent \textbf{Implications for Defense}: To be robust in this threat model, our system should avoid a single point of failure. Because the image undergoes multiple rounds of resolution scaling, and the various masks also go through steps of merging and rank-based filtering, it is difficult for low-level semantic information and noise from the original image to be propagated to subsequent models. This process thereby provides a degree of resistance against noise from adversarial attacks. Additionally, because our pipeline ultimately uses high-level semantic scores from a VLM, there is an extra layer of verification – even if segmentation produces some spurious masks, the VLM might score them low for malicious content, reducing false positives.
We note that the attacker could in theory also try to attack the VLM’s image encoder with adversarial perturbations. While possible, attacking the segmentation is generally more straightforward and yields a bigger impact (concealing entire objects). Nonetheless, using a robust image encoder (e.g., one that is adversarially trained or a Vision Transformer known for some robustness) could further harden the VLM component. In this work, we optimize adversarial perturbations specifically for SAM and Qwen-VL and subsequently use them to attack our proposed pipeline. The detailed results are presented in Section \ref{sec:Adaptive}.

\subsection{Problem Definition}
We formally define the Malicious Image Detection with Element Localization task as follows. Given an input image $I$, the goal is to determine whether $I$ violates a specific image content policy, and if so, to identify the critical elements in $I$ that led to that determination by marking their spatial extent. An “element” here refers to an object or region (which could be a part of an object) that is evidence of the malicious or disallowed activity. For example, for a drug-use event, typical elements might include drugs (pills, powders), drug paraphernalia (syringes, pipes, needles, pill bottles), and possibly the human user (if present). For violence, elements could include weapons (knives, guns), injured persons, blood, etc.

% We denote by $E_{\text{true}}(I) \subseteq E$ the set of malicious events that are actually present in image $I$ (this could be empty if the image is benign, or contain multiple event types if applicable). 
Our system should output: (1) the confidence score for a violation of a specific content safety policy, (2) for each harmful image $I \in I_{harm}$, a set of segments (masks or bounding regions) { $\hat{M}_{1}, \hat{M}_{2}, \cdots $} corresponding to the elements that contribute to the harmfulness, (3) the contribution of each segmented element in the image to the overall harmfulness of the image, presented in the form of a heatmap-like visualization $H$. Each mask $\hat{M}_{k}$ is essentially a binary map over pixels, and ideally it should closely match the ground truth mask of the object in question if one is available. The heatmap $H$ should apply a differentiated grading to all elements in the image, using clearly distinct colors to distinguish between overtly harmful, implicitly harmful, and benign elements.

Crucially, our system must handle this as a zero-shot recognition problem: we do not assume any training data that pairs images with these harmful event labels or element masks. Instead, the knowledge of what constitutes, say, a “drug-abuse event” or a “violent event” comes from the VLM’s learned common-sense and the semantic definitions (e.g., that drugs, needles are associated with drug use; guns, knives with violence). The segmentation models likewise were not trained to specifically find “malicious” objects, but generally any objects.

Success criteria: We consider the task successful if for each true harmful image, at least the key ground-truth elements are detected and segmented with reasonable accuracy. For evaluation, we define metrics in Section IV that quantify these, such as Element Detection Rate (recall and precision of ground-truth annotated elements found by the system), and Segment Success Rate (the ratio of uccessfully segmented harmful elements to the total number of harmful elements within a single image).

Our problem is closely related to open-world object detection and zero-shot segmentation, but with a specific focus on malicious or prohibited content. It is also an explainable detection problem: the system doesn’t just output a label “this image is violent” – it also provides the evidence by highlighting “this region is a gun, this region is blood”. This explainability is important for downstream decisions (e.g., a content moderator can verify the highlighted elements indeed warrant removal of the image).

In the next section, we describe our proposed pipeline designed to address this problem under the stated constraints and threat model.
\section{Methodology}
\subsection{Overview of the Proposed Pipeline}
Our method consists of a multi-stage pipeline that processes a suspicious image $I$ and produces a malicious object map highlighting regions of interest. An overview of the pipeline is as follows:

\noindent \textbf{\textcircled{1} Initial Segmentation}: Apply a segmentation model (or multiple models in parallel) to image $I$ to obtain an initial set of object masks {$M=\{m_1, m_2, \cdots, m_N\}$}. We use SAM as a primary segmenter due to its ability to generate a comprehensive set of masks covering the image. Each mask $m_i$ is a binary image of the same size as $I$, indicating the pixels belonging to one segmented region or object.

\noindent \textbf{\textcircled{2} Mask Merging and Sorting}: For initial masks $M$,we merge and choose the masks that are semantically close to enable a single mask to represent a complete, standalone semantic unit. This strategy prevents the semantic integrity of an element from being split or lost when the masks are applied.. This step ensures that each $\hat{m}_i$ contains a single, independent element. The outcome is a possibly reduced set of more independent masks {$\hat{M} = \{\hat{m}_1, \hat{m}_2, \cdots, \hat{m}_K\}$} (with $K \le N$).

\noindent \textbf{\textcircled{3} Vision-Language Scoring}: For each refined sub-mask $\hat{m}_i$, compute a maliciousness score using a vision-language model. Specifically, we generate a score $s_k$ which is 1 if the sub-object is deemed associated with any malicious event and 0 otherwise. This can be obtained by feeding the masked region (or its crop) into the VLM with an appropriate prompt (e.g., “Is this object related to a harmful or NSFW activity?") and measure the logit scores of the VLM outputs. The change in the scores can be seen as the harmful scores of the masked parts. In other words, we make full use of the VLM's ability in visual understanding by adapting a method similar to Hide-and-Seek~\cite{Hide-and-seek} and CutOut~\cite{CutOut} – how model's output changes when we erase part of the input.

\noindent \textbf{\textcircled{4} Mask Aggregation}: Using the scores as weights, fuse the sub-object masks to create a combined malicious object map $\mathcal{M}$. Essentially, masks of objects scored as malicious are merged. We assign each pixel a value reflecting how strongly it is associated with malicious elements. For example, one simple aggregation is $H(i,j) = \sum^{K}_{k=1}{tox(\hat{m}_k) \cdot \hat{m}_k(i,j)}$, i.e., using summed weights to emphasize areas with multiple flagged objects. The definition of $tox(m)$ is given in Section \ref{sec:VLMscoring}.

% \noindent \textbf{\textcircled{5} Multi-Model Integration (Ensemble)}: To improve recall and robustness, we integrate outputs from multiple segmentation backbones. We run additional segmenters (like SEEM and Grounded-SAM) on $I$ (possibly with targeted text prompts for malicious objects) and obtain their masks and maliciousness scores similarly. The final malicious map is an integration of all sources (for instance, union of masks from any model that flagged them, or an average that dilates high-confidence areas). This counters attacks that might fool one segmentation model but not another.

The above steps result in an output mask highlighting the regions containing critical elements of a detected malicious event.

In the following subsections, we explain the key components in detail, including the sub-object mask merging strategy (Section~\ref{sec:maskMerge}) and the VLM scoring mechanism (Section~\ref{sec:VLMscoring}).%, and the integration strategy against adaptive attacks (Section~\ref{sec:}). 

Figure \ref{fig:pipeline} illustrates the overall architecture of our pipeline (from input image to final highlighted output).

\subsection{Mask Merging}
\label{sec:maskMerge}

To begin, we merge the masks $M=\{m_1, m_2, \cdots, m_N\}$ generated by SAM to ensure comprehensive coverage of all distinct image regions during VLM inference. Specifically, any two masks exhibiting a high Intersection over Union (IOU $>$ 0.5) are merged. This consolidation ensures that each resulting mask corresponds to a unique element or object, thereby increasing the efficiency with which the VLM can infer and assess the toxicity level of each part.
\[
M_{\mathrm{merged}} = merge(M)
\]

One issue with off-the-shelf segmentation (especially automatic methods like SAM with no prompt) is that the resulting masks may not perfectly align with semantic “objects” of interest. SAM tends to produce masks that maximize mask quality over the image dataset and produce fine-grained segmentations with well-defined boundaries. However, when individual elements are partially occluded—for instance, a half-concealed weapon or partially covered genitalia—this detailed approach can fragment the element's semantic integrity and increase the computational cost of the subsequent VLM's inference. We therefore introduce an optimization to the SAM pipeline: a post-segmentation aggregation step. This modification is designed to preserve the semantics of individual elements to the greatest extent possible and reduce the total segment count, enabling the VLM to lock onto harmful elements more effectively. To be specific, we compute three metrics for each mask:

\begin{itemize}
    \item \textbf{Proximity of Mask Centroid to Image Center $P$.} This metric is designed to gauge the saliency of a mask. It is based on the premise that in images with malicious intent, the harmful content is typically located near the center. Consequently, we assign higher priority to masks that are more centrally located. To a specific mask $m$ of the Image $I$, $P(m)$ is given by:
    \[
        P(m)= \sqrt{(\frac{x_m^c-x_I^c}{W_I})^2+\frac{y_m^c-y_I^c}{H_I})^2}
    \]
    
    where $(x_m^c,y_m^c)$ is the coordinate of the center of the mask $m$'s bounding box, while $(x_I^c,y_I^c)$ is the coordinate of the centor of image $I$. $W_I$ and $H_I$ are the width and height of image $I$.
    \item \textbf{Area ratio of the Mask $A$.} Selecting for larger masks serves a dual purpose: it reduces the computational overhead for the subsequent VLM's inference, and it provides fewer, more substantial segments, which helps the VLM to more readily assess the harmful nature of each element. In conjunction with the proximity metric, this helps isolate the most salient elements in the image.
    \item \textbf{SAM-Generated Mask Confidence Score $C$.} This score, provided natively by the SAM model, indicates its confidence in the quality and accuracy of a mask. Objects with sharper boundaries and a consistent internal appearance tend to yield higher scores. The objective of this metric is to filter for high-quality masks that correspond to semantically coherent objects.
\end{itemize}

The product of the above three metrics will serve as the score for each mask, which is given by:

\[
    Score(m)= P(m)A(m)C(m)
\]

Then, we rank all candidate masks according to their scores and select the top-k masks. 

\[
M_{\mathrm{topk}} = \{m \in M_{\mathrm{merged}} \mid  rank_{Score}(m) \le k\}
\]

These selected masks, along with their inverted counterparts—totaling $2k$ masks—are compiled into a mask set for the next stage of VLM inference.

\[
\hat{M} = M_{\mathrm{topk}} \cup \{\bar{M_i} | M_i\in M_{\mathrm{topk}}\}
\]

where $\bar{M_i}$ stands for the inversed mask of mask $M_i$.

\subsection{Vision-Language Model Scoring}
\label{sec:VLMscoring}

After obtaining refined candidate masks, we employ a vision-language model to score each region’s relevance to malicious events. We denote the VLM as a function $F_{\text{VLM}}(I, m)$ that takes the image $I$ and a mask $\hat{m}$ and outputs a logit tensor. 

For each candidate mask generated in the previous step, we present the VLM with a version of the original image where the region defined by that mask has been blacked out. By observing the resulting change in the VLM's output, we can infer the importance of the occluded region.

Specifically, we query the VLM with a carefully constructed prompt, such as, "Analyze the following image. Does this image contain harmful content? Please answer with 'Yes' or 'No'." We adopt a policy-as-prompt approach~\cite{palla2025policy}, incorporating the content policy directly into the prompt to help the model better assess the harmfulness of an image. The logit value for the token 'Yes' is then treated as the toxicity score for the given input. The contribution of a single mask (m) is quantified by the drop in this score when its corresponding region is hidden:

\[
logits = F_{\text{VLM}}(I, m)
\]

where $I$ is the original image and $I_m$ is the image with mask $m$ occluded. We assume binary for simplicity: 1 = harmful element, 0 = benign. We extract the raw output score (logit) from VLM corresponding to the "harmful" class ($logit_1$) and "harmless" class ($logit_0$). The toxicity score of the image with mask $m$ occluded is given by:
\[
    tox(m) =\frac{e^{logit_1}}{e^{logit_1}+e^{logit_0}}
\]

\subsection{Toxicity Heatmap Construction}
Finally, to create a comprehensive toxicity distribution map for the entire image, we overlay all candidate masks, weighting each one by its calculated score $tox(m)$. This produces a heatmap where the intensity of each pixel corresponds to the VLM's judgment of its toxicity, providing a clear visual representation of the model's reasoning.

First, we create an initial map $H$ by performing a weighted aggregation of all masks at each pixel, using their respective toxicity scores $tox$ as the weights. 
\[
    H(i,j) = \sum^{K}_{k=1}{tox(\hat{m}_k) \cdot \hat{m}_k(i,j)}
\]
where $K$ is the total number of masks $\hat{M}$ and $H(i,j)$ is the initial aggregated score at pixel coordinates $(i,j)$.

Subsequently, this aggregated map undergoes min-max normalization to scale the final distribution to a $[0,1]$ range.
\[
    \mathcal{M}(i,j) = \frac{H(i.j) - min(H)}{max(H) - min(H) +\epsilon} \ tox(I)
\]
where $min(H)$ and $max(H)$ are the minimum and maximum values found across the entire heatmap $H$. $\epsilon$ is a small constant (e.g., 1e$-$8) added to the denominator to prevent division by zero in cases where all pixel values in $H$ are identical. $tox(I)$ is the toxicity score of the original unmasked image.

The resulting heatmap, $\mathcal{M}$ , represents the final toxicity distribution, illustrating the toxicity value of each relatively independent element from the VLM's perspective. Through this, we have achieved an element-level toxicity localization for images that is based on the VLM's deep understanding and features high granularity and robustness to interference.
\section{Experimental Settings}
\subsection{Dataset and Annotations}
\label{sec:dataset}
We evaluate our method on a curated NSFW-Malicious Image Dataset. This dataset consists of images gathered from both open-source NSFW collections and AI generated contents (specifically from prompting FLUX~\cite{flux}).We focused on four primary malicious event categories: Illegal Activities, Sexual \& Pornography, Violence \& Horror, and Discrimination \& Extremism. To provide finer-grained coverage, our dataset further includes eleven representative subcategories commonly regulated by content safety policies: gambling, theft, robbery, drug use, firearms, knives, gore and horror, self-harm, pornography, discrimination, and extremism. The dataset contains a total of 790 images, spanning a wide range of real-world and synthetic scenarios. For example, illegal images include people consuming drugs with powders on tables, people robbing banks, knife in hands, etc.; violent images include armed assaults, fights, scenes with visual horror and blood, etc.; porn images include pornographic scenes and explicit nudity; Discriminate and extremism images include religious blasphemy and extremism marks.

Each image in the dataset was manually annotated with element-level masks. Our annotators drew polygons around every object that is critical to the malicious interpretation of the image. On average, each image has 1–5 such elements annotated. For instance, in a drug image, masks might include the syringe, the lighter, and the substance being prepared; in a violent image, the weapons and possibly wounds; in a porn image, exposed body parts (genitals) or sex toys. We also annotated the category of each element (choosing from a predefined list per category, e.g., “syringe”, “gun”, “knife”, “male genitalia”, etc.). These granular annotations allow us to quantitatively measure if the method detects each element and how well the predicted mask overlaps it.

It’s worth mentioning that existing public datasets only partially fulfill our needs. OpenImages\cite{OpenImages} and COCO\cite{coco} have some labels for knives, guns, etc., but not comprehensive or focused on illicit contexts. The Hateful Memes dataset\cite{Memes} involves multimodal content (image + text) but not exactly what we need. A specialized dataset called DRUNalia\cite{zhao2020drunaliacap} was introduced for captioning drug paraphernalia, which has 20 categories of drug-related items. We drew inspiration from it and generated similar images in our drug category. For violence, we leveraged some images from VHD11k~\cite{VHD11k} and added more from FLUX generation. The porn content images are from nsfwdataset on github~\cite{nsfwdataset}, filtered to those with explicit intercourse or nudity.

We emphasize that no model is trained or fine-tuned on these images – they are purely for evaluation. Importantly, our method leverages the native capabilities of VLMs to localize harmful elements, allowing the underlying content safety policy and image categories to be flexibly adapted without any need for model retraining.

\subsection{Baseline Methods}
We compare our proposed method against several baselines to highlight the advantages of each component. The baselines include:

% \noindent \textbf{Image-level NSFW Classification}: Though not a focus on localization, we use a standard NSFW detector (based on a ResNet or Vision Transformer fine-tuned for NSFW categories) to see if it correctly flags the images. This baseline yields an image-level yes/no and perhaps category, but no localization. It helps gauge how challenging the images are for a generic classifier.

% \noindent \textbf{LLMSeg}:LLMSeg~\cite{wang2024llmseg} is a reasoning segmentation model that combines SAM~\cite{kirillov2023segment} and LLaVA~\cite{llava}. first segments an image with SAM and then presents the resulting segments as options to LLaVA for precise object localization. It was originally used to precisely locate specific objects in an image; we modified it to locate harmful elements instead.
\noindent \textbf{LISA}: LISA~\cite{lai2024lisa} is a multimodal model designed to perform a task called "reasoning segmentation." The goal of reasoning segmentation is to generate a precise segmentation mask based on complex, conversational, or implicit user instructions. Instead of a simple command like "segment the car," a user might ask something that requires reasoning and world knowledge, such as, "Can you show me the part of the image that is most suitable for a picnic?" In this section, we feed our content safety policy to LISA, instructing it to segment the harmful regions within the image. This baseline demonstrates the transfer capability of a non-zero-shot method to the field of harmful content localization, for comparison with our zero-shot approach.

\noindent \textbf{Grad-CAM on VLM}: Grad-CAM~\cite{selvaraju2017gradcam} is a popular explainability technique that produces coarse localization by gradient back-projection. We attempt Grad-CAM on a VLM classifier – e.g., ask Qwen-VL if the image is harmful (yes/no) and use Grad-CAM on its transformer layers to see what pixels most affect the outputs. This baseline provides a localization of malicious content without explicitly segmenting objects.

\noindent \textbf{Attention Maps}: Attention maps are a native explainability feature of Transformer-based models. They reveal the model's focus by visualizing the self-attention weights between input tokens, indicating which image regions were most influential in its decision-making process. We apply this concept by prompting a VLM like Qwen-VL with a harmfulness query (yes/no), and then we extract and aggregate the attention scores from its transformer layers, particularly the attention paid by the classification token to the various image patches. This approach similarly yields a heatmap localizing potentially malicious content, leveraging the model's intrinsic architecture rather than relying on gradient information.

\noindent \textbf{Qwen-VL Zero-Shot Referring}: We utilize the VLM’s referring expression capability to directly query for malicious elements. For instance, we prompt Qwen-VL with: “Locate the drugs in the image” or “Show the bounding box of any weapons in this image.” If the model is able to output bounding boxes, this serves as a zero-shot detection baseline. We attempt multiple prompts per category and count it as a detection if the box overlaps a ground truth element. This tests the VLM’s inherent detection ability without our pipeline.

% \noindent \textbf{SAM with Manual Labeling}: This baseline simulates using SAM to segment an image and then manually (or via a heuristic) deciding which masks are malicious. Concretely, we run SAM to get all masks, then for fairness we use ground-truth knowledge to pick which of SAM’s masks correspond to the annotated malicious elements. This isn’t a practical method (it uses ground truth), but it gives an upper bound on how well SAM’s raw masks cover the true objects. We measure the average IoU of the best-matching SAM mask to each ground-truth element.

% \noindent \textbf{Grounded-SAM (Text Prompted Segmentation)}: Here we use the known category labels of elements as prompts. For a given image known to contain, say, “gun” and “knife,” we prompt Grounded-SAM with those words to get masks. In evaluation, we of course supply the full list of possible relevant words (drugs: “pills, syringe, bottle, powder, spoon, etc.”; violence: “gun, knife, sword, blood,” etc.; sexual: “breast, penis, condom,” etc.) to simulate an oracle that knows what to look for. This baseline tests how effective open-vocabulary segmentation is with perfect prompt knowledge. It uses the implementation from Ren et al. with the same Grounding DINO checkpoint as in their paper.

% \noindent \textbf{SEEM (Text Prompted)}: Similarly, we evaluate SEEM by giving it the text prompts of relevant objects (one by one) and obtaining masks. SEEM can return multiple masks for a single text query if multiple instances exist, which is helpful. We then union all masks obtained from all prompts for that image.

\noindent \textbf{Our Method Variants}: We also compare variants of our approach to isolate contributions: (i) Ours-no-merge: the pipeline without the mask merging and sorting step (i.e., using SAM’s original masks directly for VLM scoring); (ii) Ours-no-score: instead of merging sub masks by VLM toxicity score, treat each mask independently and see if at least one mask overlaps each ground truth (this checks if the fusion and weighting improve localization tightness).

For each baseline that produces masks or boxes, we evaluate them on the same metrics described below.

\subsection{Evaluation Metrics}
We use the following metrics to quantify performance:

% \noindent \textbf{Event Detection Rate}: This measures the ability to correctly identify the presence of a malicious event in the image. Since each image in our test set has a ground truth label (Drug, Sex, Violence), EDR is essentially the classification accuracy for each category (or we report recall at high precision, since false positives are also bad). We count an image as a true positive if the method outputs the correct event label for it. If multiple events can be output, we treat each category separately (like multilabel detection).

\noindent \textbf{Element Detection Rate}: For each ground-truth annotated element (mask) in the image, we check if the method found a corresponding predicted mask. Because our method outputs a combined mask highlight rather than discrete labeled masks per object, we consider an element “detected” if at least 50\% of its area (mask) is covered by the predicted malicious map $M_{\text{fused}}$ (for segmentation-based methods). For methods that output boxes, we convert ground truth mask to its bounding box and check IoU with predicted box $>$ 0.5. We then compute the Recall: the percentage of ground-truth elements that were successfully detected. We also measure the Precision (percentage of predicted regions that correspond to true malicious elements) to gauge false positives. We do not compute Accuracy because True Negatives are meaningless in the context of our task.

% \noindent \textbf{Element Boundary Coincidence (IoU)}: If an element is detected, how precise is the localization? We calculate the Intersection over Union (IoU) between the predicted mask region and the ground truth mask for each element. For methods that don’t produce pixel masks (like Grad-CAM heatmaps or VLM boxes), we can compute a proxy IoU (e.g., by thresholding heatmap or using bounding box overlap). We report the mean IoU over all detected elements (and perhaps the distribution quantiles).

\noindent \textbf{Segment Success Rate}: To quantify how well the harmful elements are extracted in our method, human experts inspect each image to verify the segmentation of harmful elements. It is important to note that the evaluation focuses on \textit{whether} an element is cleanly partitioned by a distinct boundary, not on its identification as harmful. An element is deemed 'successfully segmented' if its boundary is clear and does not overlap with other elements. The Segment Success Rate is defined as the ratio of successfully segmented harmful elements to the total number of harmful elements within a single image.

\noindent \textbf{Bounding Box Type}: Since VLM zero-shot queries can only produce bounding boxes, which differ from the fine-grained masks obtained by our method, we proposed a four-type rubric to better evaluate the quality of these bounding boxes. Each image was assessed by human experts according to the following scenarios:
\begin{itemize}
    \item Type 1: Completely Incorrect. In this scenario, the bounding box highlights a harmless region, and the actual harmful content is not marked at all.
    \item Type 2: Partial Omission. In this scenario, the bounding box correctly marks some of the harmful content but omits other harmful regions.
    \item Type 3: Excessive Bounding. In this scenario, the bounding box covers all harmful regions, but its area is so large (over 80\% of the image) that the localization is rendered practically meaningless.
    \item Type 4: Perfect Selection. In this scenario, the bounding box accurately and tightly frames all harmful regions with a precise, minimal area.
\end{itemize}

\noindent \textbf{False Positive Rate}:Since benign images do not contain harmful elements, it is not possible to calculate the Precision and Recall for Element Detection. Therefore, we adopt the false positive rate (FPR) as the metric for benign images. A false positive ($FP$) occurs if our method detects a benign image as harmful (score $>$ 0.5). The FPR is then calculated as:

\[
FPR = \frac{FP}{ N_{benign}}
\]

\noindent where $N_{benign}$ is the total number of benign images.

\noindent \textbf{Robustness under Attack}: We craft a small subset of images with adversarial perturbations targeting SAM\cite{kirillov2023segment} (following the approach of Attack-SAM\cite{zhang2023attacksam} for mask removal) and test whether the methods still detect the elements. We measure the drop in Element Detection Rate under this attack for SAM-alone vs our integrated method.
Additionally, we qualitatively assess the quality of explanations: i.e., does the highlighted region align well with what a human would consider the problematic content? This is anecdotal but important for practical adoption.

All metrics are calculated and evaluated by human experts based on a random sample of 100 images in all categories. All metrics are computed per category as well (since performance may differ: e.g., porn content might be easier to detect than concealed drugs). In the following section, we present the results along these metrics.

\section{Experimental Results and Analysis}
\subsection{Quantitative Results}
\begin{table}[!ht]
    \centering
    \caption{Our method on Qwen2.5VL-7B.}
    \label{tab:ours}
    \resizebox{0.5\textwidth}{!}{
    \begin{tabular}{l|c|c|c|c|c}
    \toprule
        Metrics & Illegal & Porn & Violence & Extremism & Overall  \\ 
    \midrule
        Segment Success Rate & 0.8672 & 0.9610 & 0.9136 & 1.0000 & 0.9205  \\ 
        Recall & 0.8028 & 0.9067 & 0.8182 & 1.0000 & 0.8582  \\ 
        Precision & 0.7037 & 0.8710 & 0.7312 & 0.8780 & 0.7805  \\ 
    \bottomrule
    \end{tabular}
    }
\end{table}
\begin{table}[!ht]
    \centering
    \caption{Direct response bounding boxes on Qwen2.5VL-7B.}
    \label{tab:direct}
    \resizebox{0.5\textwidth}{!}{
    \begin{tabular}{l|c|c|c|c|c}
    \toprule
        Metrics & Illegal & Porn & Violence & Extremism & Overall  \\ 
    \midrule
        Recall & 0.5179 & 0.6438 & 0.5821 & 0.8333 & 0.5941  \\ 
        Precision & 1.0000 & 0.9592 & 1.0000 & 1.0000 & 0.9836  \\ 
    \bottomrule
    \end{tabular}
    }
\end{table}
\begin{table}[!ht]
    \centering
    \caption{Direct response bounding box type on Qwen2.5VL-7B.}
    \label{tab:direct-types}
    \resizebox{0.5\textwidth}{!}{
    \begin{tabular}{l|c|c|c|c|c}
    \toprule
        Scenario Type               & Illegal   & Porn   & Violence  & Extremism & Overall  \\ 
    \midrule
        Complete Incorrect          & 4\%       & 3\%           & 0\%       & 20\%      & 6\%    \\ 
        Partial Omission            & 72\%      & 33\%          & 62\%      & 0\%       & 51\%   \\
        Excessive Bounding          & 0\%       & 24\%          & 25\%      & 40\%      & 7\%    \\
        Perfect Selection           & 24\%      & 39\%          & 23\%      & 40\%      & 30\%   \\
    \bottomrule
    \end{tabular}
    }
\end{table}
\begin{table}[!ht]
    \centering
    \caption{LISA on ourdataset (LISA-13B-llama2-v1).}
    \label{tab:LISA}
    \resizebox{0.5\textwidth}{!}{
    \begin{tabular}{l|c|c|c|c|c}
    \toprule
        Metrics & Illegal & Porn & Violence & Extremism & Overall  \\ 
    \midrule
        Segment Success Rate & 0.4796 & 0.5737 & 0.5608 & 0.6000 & 0.5418  \\ 
        Recall & 0.4722 & 0.6500 & 0.5432 & 0.5000 & 0.5479  \\ 
        Precision & 0.8718 & 0.9512 & 0.8148 & 0.7500 & 0.8696  \\ 
    \bottomrule
    \end{tabular}
    }
\end{table}

% TODO

\noindent \textbf{Overall Performance}: Table \ref{tab:ours} and Table \ref{tab:direct} show the performance of our method and VLM zero-shot query. Our method achieves a Element Detection Recall of 85.8\% across all categories, siginificantly higher than the VLM zero-shot queries (58.4\%). This means the vast majority of ground-truth annotated objects were successfully highlighted, missing very few. In terms of Element Detection Precision, we reach 78.0\%, lower than the VLM zero-shot queries (98.3\%). Additionally, to illustrate the inherent limitations of the direct query approach, we classified the bounding boxes (bboxes) returned by the VLM into four categories. As shown in Table \ref{tab:direct-types}, a mere 30\% of the total bboxes precisely and accurately framed all harmful elements.

For benign images, we used the ImageNet ILSVRC2012~\cite{ILSVRC15} dataset as our benign dataset and randomly sampled 100 images for manual verification. With our method, we achieved a False Positive Rate (FPR) of 1\%. Through our verification, the only false positive case in the sample was an image of a pistol. Although it did not depict an illegal scene, such as pointing the gun at a person, the firearm itself does carry a certain degree of harmful semantics.

\noindent \textbf{Category-wise Breakdown}: The performance is highest on the extremism category (recall 100\%, precision 87.8\%). This is likely because objects related to extremism (such as flags and symbols) tend to have well-defined boundaries and distinct visual features that set them apart from other elements. For porn images, recall and precision is lower (90.7\% and 87.1\%) - failures here often involved borderline cases (e.g., partial nudity not clearly porn context). For illegal activities and violence, recall is around 80\% and precision is around 70\%. In these scenarios, harmful elements may not appear as a specific object (such as casino scenes and blood stain).

\noindent \textbf{Baselines Comparison}: 
We input our content safety policy into LISA~\cite{lai2024lisa} and instruct it to select the harmful content regions in the image. LISA did not perform well on the harmful content localization task in this paper. Although it achieved a high Precision (86.96\%), its Recall was only 54.79\%. Because LISA is designed to output only a single \verb|SEG| token~\cite{lai2024lisa}, it has a weak capability for handling multiple elements, often selecting only one harmful element when several are present in an image.

The zero-shot VLM queries achieves higher precision in all categories among all methods. Qwen-VL\cite{bai2025qwen} tends to identify and output a region that it considers definitively harmful. The trade-off, however, is that this approach fails to comprehensively identify all harmful elements within the image. Furthermore, across all tested harmful images, Qwen-VL only designated a harmful region in 89\% of cases; in the remaining images, it did not recognize any harmful areas at all.

Additionally, it is important to note that not all VLMs possess Qwen-VL's capability to output accurate bounding boxes. We tested similar queries on both Llava-1.5\cite{llava} and Meta-Llama 3.2 Vision\cite{llama}. In our tests, Llava-1.5 invariably bounded the entire image without exception, while Meta-Llama explained that it could not provide such a functionality.

Grad-CAM is a popular explainability method used to understand a model's internal assessment of each part of an image. Although Grad-CAM was first applied to CNN-based models, it can also be used to visualize decisions in Vision-Language Models (VLMs), since their internal Vision Transformers (ViTs)\cite{vit} likewise preserve the structural information of the image.

\begin{figure}[!ht]
    \centering
    \begin{minipage}[t]{0.14\textwidth}
        \centering
        \includegraphics[width=\linewidth]{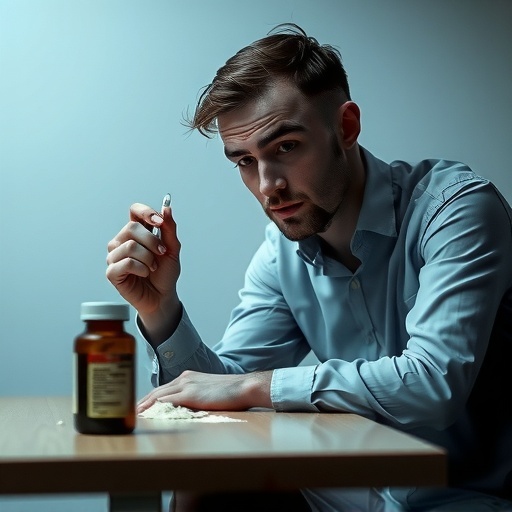}
        \vspace{2pt}
        \textbf{(a)} Original Image
    \end{minipage}
    \begin{minipage}[t]{0.14\textwidth}
        \centering
        \includegraphics[width=\linewidth]{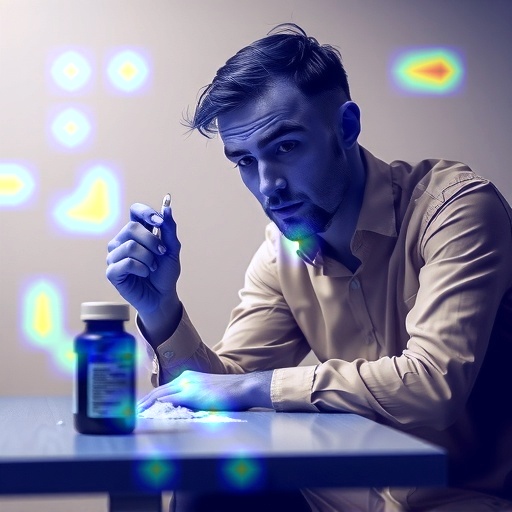}
        \vspace{2pt}
        \textbf{(b)} Grad-CAM Visualization
    \end{minipage}
    \begin{minipage}[t]{0.14\textwidth}
        \centering
        \includegraphics[width=\linewidth]{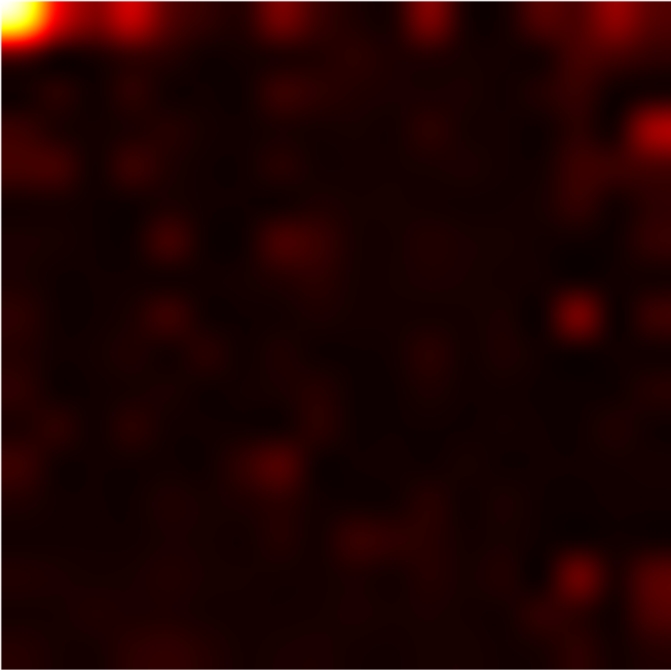}
        
        \vspace{2pt}
        \textbf{(c)} Attention Map
    \end{minipage}
    \caption{Explainability Methods on Qwen2.5VL-7B (All Transformer Layers Overlay)}
    \label{fig:Grad-CAM-and-attention}
\end{figure}

We attempted to apply Grad-CAM at the end of each transformer layer and found that it only offered a degree of interpretability in the shallower layers. This aligns with research by Zhang et al.\cite{redundancy}, which also suggests that VLMs primarily process visual information streams in their early layers. As shown in Figure \ref{fig:Grad-CAM-and-attention}(b), we ultimately found that the heatmaps provided by the Grad-CAM method contain considerable noise; in other words, Grad-CAM focuses on an excessive number of harmless regions. For an abstract and challenging task like identifying harmful content, Grad-CAM could not effectively or accurately pinpoint the harmful regions.

We also experimented with using \textbf{attention maps} to visualize the model's decision-making. Attention weights are often considered a method for providing explainable insights~\cite{wiegreffe2019attention}. We attempted to extract the attention weights of the final token with respect to all visual tokens from every transformer layer, which we then aggregated and remapped back into a two-dimensional format. However, our results indicated that, in the case of Qwen, these attention weights do not offer meaningful interpretability, which is shown in Figure \ref{fig:Grad-CAM-and-attention}(c).

\subsection{Qualitative Analysis}

\begin{figure}[!ht]
    \centering
    \begin{minipage}[t]{0.15\textwidth}
        \centering
        \includegraphics[width=\linewidth]{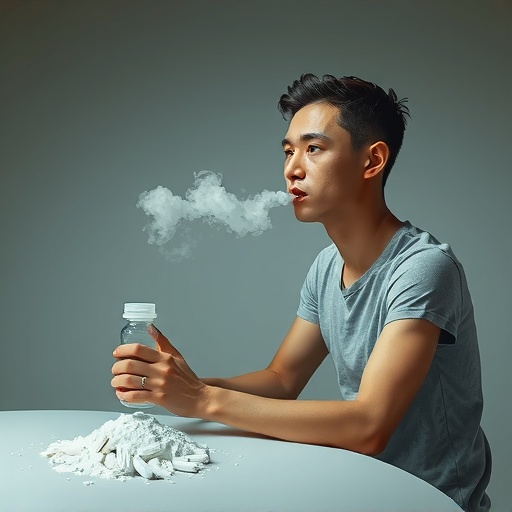}
        \vspace{2pt}
        \textbf{(a)} Original Harmful Image
    \end{minipage}
    \hfill
    \begin{minipage}[t]{0.17\textwidth}
        \centering
        \includegraphics[width=\linewidth]{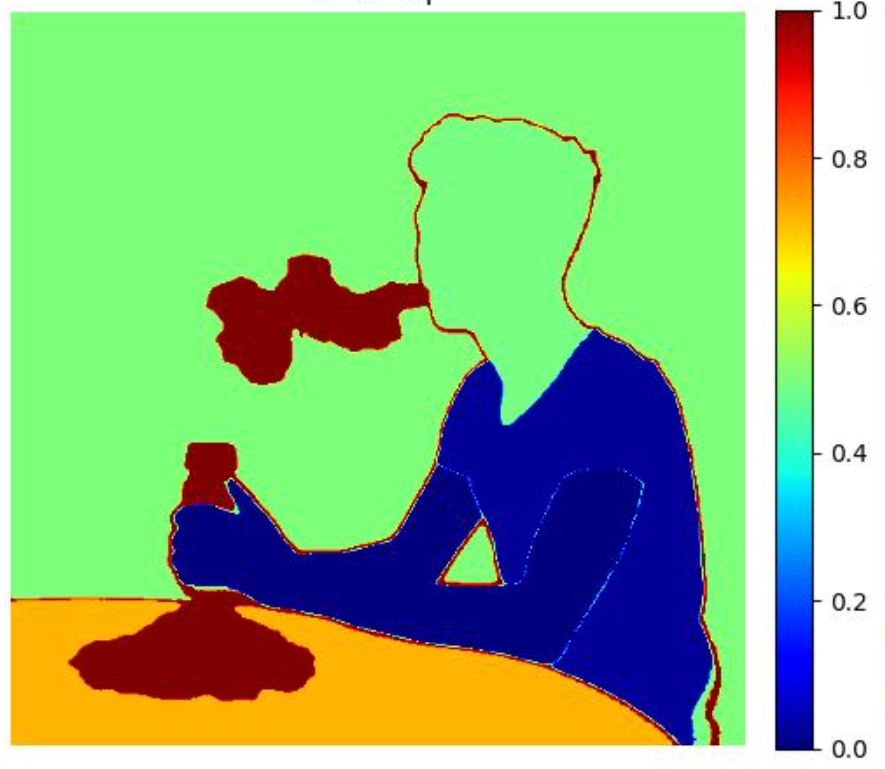}
        \vspace{2pt}
        \textbf{(b)} Our Method Heatmap
    \end{minipage}
    \hfill
    \begin{minipage}[t]{0.15\textwidth}
        \centering
        \includegraphics[width=\linewidth]{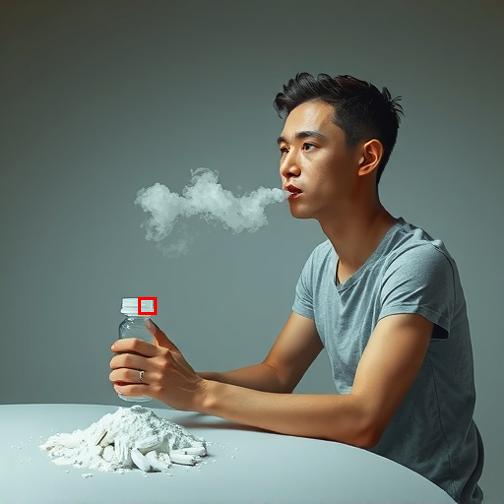}
        
        \vspace{2pt}
        \textbf{(c)} Direct Query Bounding Box
    \end{minipage}
    \caption{Drug image comparison of our method with direct query (Qwen2.5VL-7B)}
    \label{fig:drug_compare}
\end{figure}

We present several representative examples to illustrate the behavior of our system:

\noindent \textbf{Illegal Scene - Drug Abuse}: Figure \ref{fig:drug_compare}(a) shows a harmful image depicting a man holding a pill bottle and exhaling smoke, with suspicious powder on the table. Our method segments out the powder, the bottle, and even the smoke in the air. All three are highlighted clearly in the output mask. The VLM correctly and hierarchically assigned toxicity scores to these elements, ultimately constructing our toxicity heatmap. As shown in Figure \ref{fig:drug_compare}(b), within this map, the 'powder', 'pill bottle', 'smoke', and the silhouette of the person were labeled as the most harmful parts, while the remaining areas were assigned clearly stratified heat values based on their degree of contextual semantic relevance. However in Figure \ref{fig:drug_compare}(c), direct query get a bounding box only focus on part of the pill bottle, turning a blind eye to 'powder' and 'smoke', the other two harmful elements.

\begin{figure}[!ht]
    \centering
    \begin{minipage}[t]{0.15\textwidth}
        \centering
        \includegraphics[width=\linewidth]{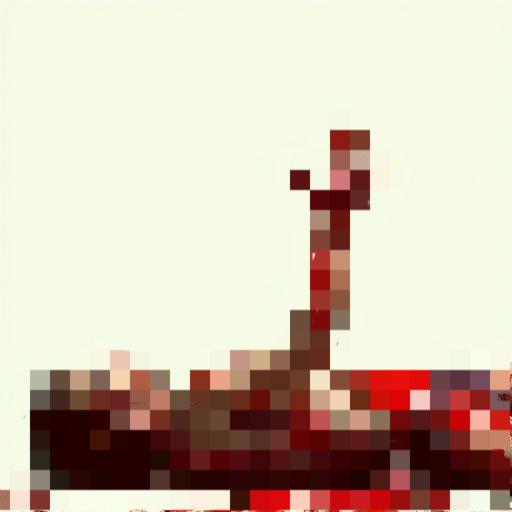}
        \vspace{2pt}
        \textbf{(a)} Original Harmful Image
    \end{minipage}
    \hfill
    \begin{minipage}[t]{0.17\textwidth}
        \centering
        \includegraphics[width=\linewidth]{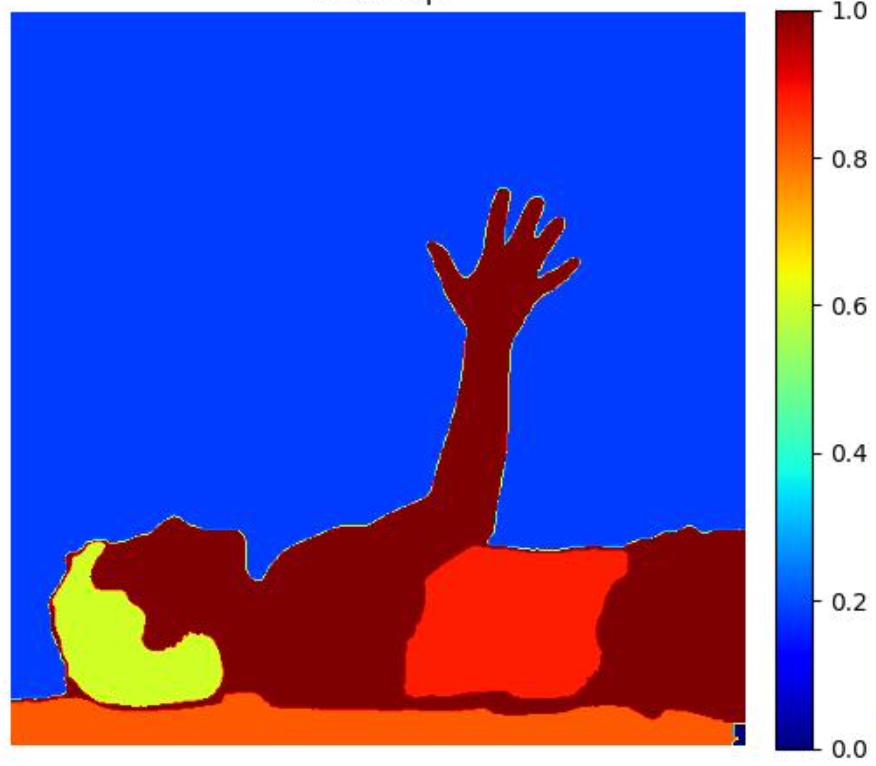}
        \vspace{2pt}
        \textbf{(b)} Our Method Heatmap
    \end{minipage}
    \hfill
    \begin{minipage}[t]{0.15\textwidth}
        \centering
        \includegraphics[width=\linewidth]{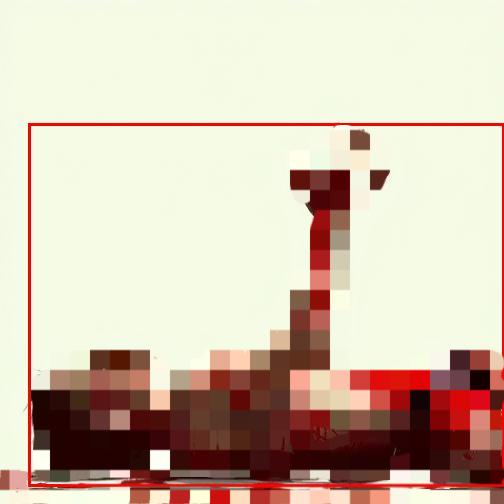}
        
        \vspace{2pt}
        \textbf{(c)} Direct Query Bounding Box
    \end{minipage}
    \caption{Gory image comparison of our method with direct query (Qwen2.5VL-7B)}
    \label{fig:violent_compare}
\end{figure}

\noindent \textbf{Violent Scene}: Figure \ref{fig:violent_compare}(a) shows a harmful image depicting a body lying on the ground with blood everywhere. As shown in Figure \ref{fig:violent_compare}(b), our method segments out the lying body and the blood-stained ground. In Figure \ref{fig:violent_compare}(c), direct query from Qwen2.5VL gets a bounding box similar to the high heat region. However, this bounding box covers such a large area of the image that it fails to precisely pinpoint the location and contours of the harmful elements. Furthermore, a portion of the "blood-stained ground" was not included within the box. Consequently, its ability to localize harmful elements is inferior to our method.

\begin{figure}[!ht]
    \centering
    \begin{minipage}[t]{0.15\textwidth}
        \centering
        \includegraphics[width=\linewidth]{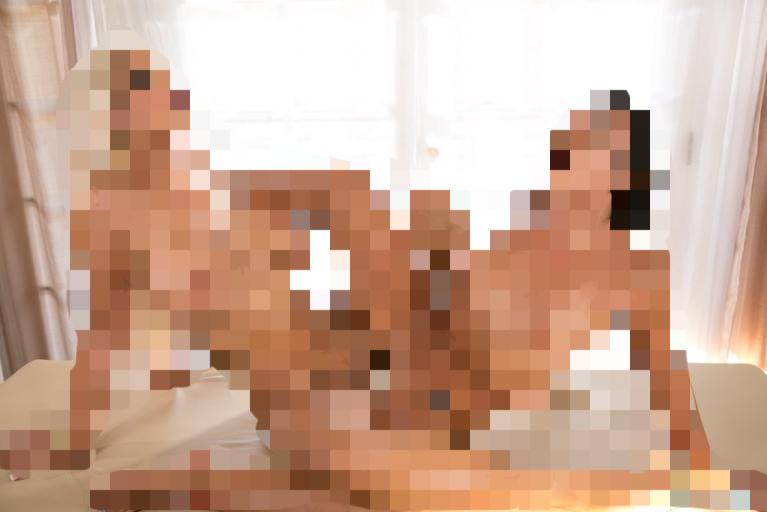}
        \vspace{2pt}
        \textbf{(a)} Original Harmful Image
    \end{minipage}
    \hfill
    \begin{minipage}[t]{0.17\textwidth}
        \centering
        \includegraphics[width=\linewidth]{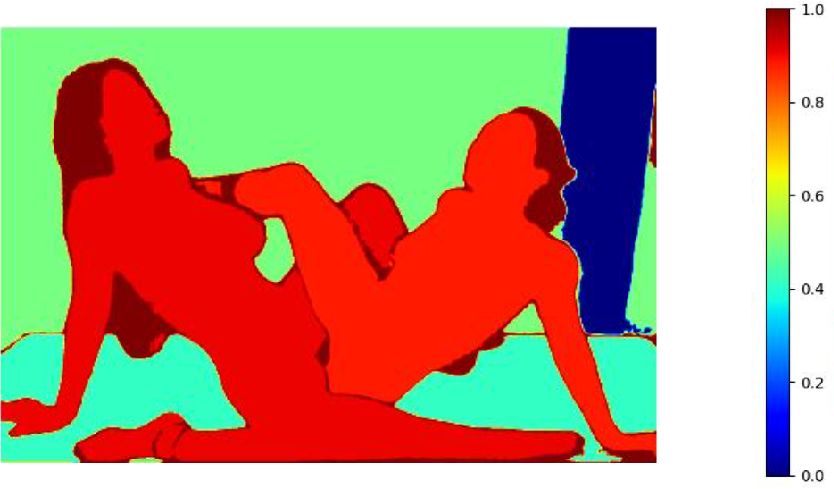}
        \vspace{2pt}
        \textbf{(b)} Our Method Heatmap
    \end{minipage}
    \hfill
    \begin{minipage}[t]{0.15\textwidth}
        \centering
        \includegraphics[width=\linewidth]{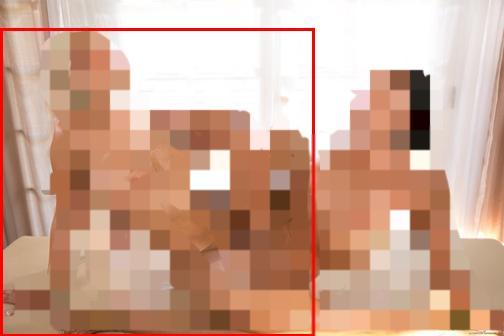}
        \vspace{2pt}
        \textbf{(c)} Direct Query Bounding Box
    \end{minipage}
    \caption{Porn image comparison of our method with direct query (Qwen2.5VL-7B)}
    \label{fig:porn_compare}
\end{figure}

\noindent \textbf{Porn Content}: Figure \ref{fig:porn_compare}(a) shows an image with two partially clothed individuals engaged in a sexual act. Our method highlighted the exposed body parts which correspond to the explicit content. The VLM in this case was triggered by recognizing the nudity and thus our method generated a toxicity heatmap with precise segmentation and a logical toxicity distribution (Figure \ref{fig:porn_compare}(b)). Notably, the hair of the two individuals in the image was incorrectly marked as having the highest toxicity value; this is likely because occluding the hair during the masking process damaged the original semantics of the image. For comparison, the bounding box obtained by directly querying the VLM is shown in Figure \ref{fig:porn_compare}(c). The VLM only marked one of the nude individuals, completely ignoring the other.. 

\subsection{Evaluation Under Adaptive Attacks}
\label{sec:Adaptive}

\subsubsection{Attack against Segment Anything Model (SAM)}
Methods for attacking SAM already exist~\cite{zhang2023attacksam,GRAT}. These techniques allow us to add adversarial noise to an original image, making it difficult for SAM to process the perturbed image correctly and causing it to output lower-quality masks. We further analyze how the integration strategy fares against adversarial attempts. For our approach, this type of attack could potentially degrade the quality of our sub-images, leading to a final toxicity heatmap that cannot clearly delineate the contours of the harmful elements. In practice, we adopt PGD (Projected Gradient Descent) to generate adversarial images.

% \begin{algorithm}
% \caption{PGD Attack on SAM Encoder}
% \label{alg:optimized_pgd}
% \begin{algorithmic}[1]
%     \STATE \textbf{Input:}
%     \STATE \quad $M_{enc}$: SAM model with image encoder
%     \STATE \quad $x$: Original input image tensor, normalized to $[0, 1]$
%     \STATE \quad $N$: Number of attack iterations
%     \STATE \quad $\epsilon$: Maximum $L_\infty$ perturbation norm
%     \STATE \quad $\alpha$: Attack step size (default $1/255$)
%     \STATE \textbf{Output:}
%     \STATE \quad $x_{adv}$: Adversarial image
%     % \vspace{0.5em}
%     \STATE $f_{orig} \gets M_{enc}(x)$
    
%     % \vspace{1em}
    
%     \STATE $\delta \gets \text{Uniform}(-\epsilon, \epsilon)$ 
%     \STATE $g \gets 0$ 
    
%     % \vspace{1em}
    
%     \FOR{$i = 0$ to $N-1$}
%         \STATE $x_{pert} \gets x + \delta$
%         % \vspace{0.5em}
%         \STATE $f_{pert} \gets M_{enc}(x_{pert})$
%         \STATE $\mathcal{L} \gets -\text{MSE}(f_{pert}, f_{orig})$ 
        
%         % \vspace{0.5em}
%         \STATE $g \gets \nabla_{\delta} \mathcal{L}$
        
%         % \vspace{0.5em}
%         \STATE $\delta \gets \delta - \alpha \cdot \text{sign}(g)$
        
%         % \vspace{0.5em}
%         \STATE $\delta \gets \text{clamp}(\delta, -\epsilon, \epsilon)$
        
%         % \vspace{0.5em}
%         \STATE $x_{adv} \gets \text{clamp}(x + \delta, 0, 1)$
        
%         % \vspace{0.5em}
%         \STATE $\delta \gets x_{adv} - x$
%     \ENDFOR
    
%     % \vspace{1em}
    
%     \RETURN $x_{adv}$
% \end{algorithmic}
% \end{algorithm}

% 简化表达的伪代码
\begin{algorithm}
\caption{PGD Attack on SAM Encoder}
\label{alg:optimized_pgd}
\begin{algorithmic}[1]
\STATE \textbf{Input:} Image $x \in [0,1]$, SAM encoder $M_{enc}$, steps $N$, perturb limit $\epsilon$, step size $\alpha$
\STATE \textbf{Initialize:} $f_{orig} \gets M_{enc}(x)$, $\delta \sim \mathcal{U}(-\epsilon, \epsilon)$
\FOR{$i = 1$ to $N$}
    \STATE $f_{pert} \gets M_{enc}(x + \delta)$
    \STATE $\delta \gets \text{clip}(\delta - \alpha \cdot \text{sign}(\nabla_\delta \text{MSE}(f_{pert}, f_{orig})), -\epsilon, \epsilon)$
    \STATE $\delta \gets \text{clip}(x + \delta, 0, 1) - x$
\ENDFOR
\STATE \textbf{Return:} $x_{adv} \gets x + \delta$
\end{algorithmic}
\end{algorithm}

Using random chosen 100 images where we knew the target objects, we generated adversarial examples against SAM. In detail, we adopt PGD algorithm described in Algorithm~\ref{alg:optimized_pgd} to generate adversarial images. Each adversarial example aimed to make the respective model fail to segment the target object (mask removal objective). In particular, we set $\epsilon=8/255$ to strike a balance between preserving the original semantics of the image and maximizing the strength of the adversarial noise.

The loss function $\mathcal{L}$ for the PGD attack is defined as the negative Mean Squared Error (MSE) between the feature embeddings of the original image and the perturbed image. The goal is to maximize the distance between these embeddings, which is equivalent to minimizing the negative MSE.

Let $M_{enc}$ be the SAM image encoder.
Let $x$ be the original image tensor.
Let $x_{pert}$ be the perturbed image tensor, where $x_{pert} = x + \delta$.

The feature embedding for the original image is:
\[
f_{orig} = M_{enc}(x)
\]

The feature embedding for the perturbed image is:
\[
f_{pert} = M_{enc}(x_{pert})
\]

The loss function $\mathcal{L}(x_{pert}, x)$ is then formulated as:
\[
\mathcal{L}(x_{pert}, x) = - \text{MSE}(f_{pert}, f_{orig}) 
\]

where $D$ is the total number of elements in the feature embedding vector. By minimizing this loss, the optimization process actively pushes the perturbed embedding $f_{pert}$ away from the original embedding $f_{orig}$.

\begin{figure}[!ht] 
\centering 
\includegraphics[width=0.48\textwidth]{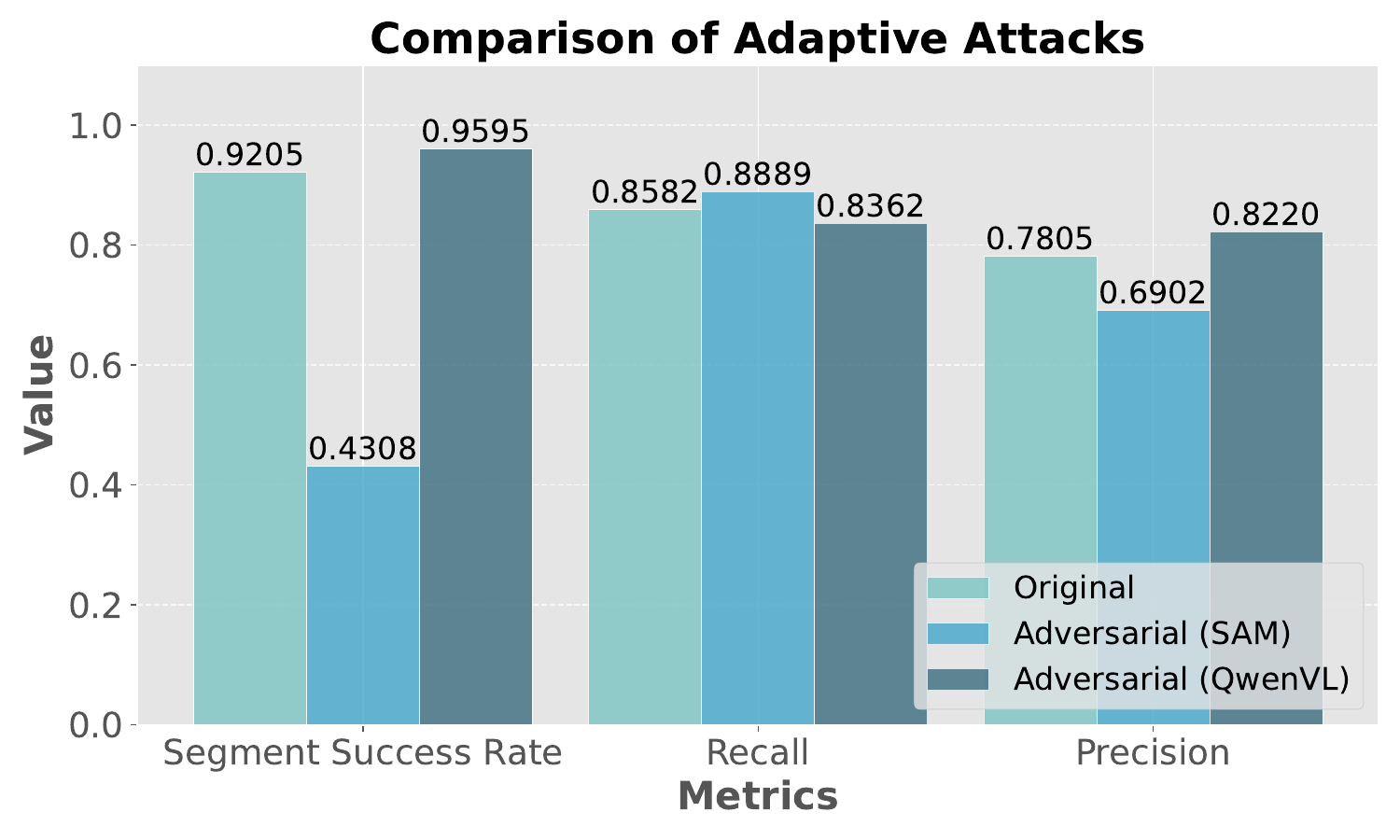}
\caption{Comparison of original harmful images vs. adversarial images against SAM vs. adversarial images against Qwen2.5VL on our method.} 
\label{fig:adv_compare} 
\end{figure}

We then ran our full pipeline on these. As shown in Figure \ref{fig:adv_compare}, the Segment Success Rate declined by 48.97\%. The PGD attack is very effective against Segment Anything Model. However, the Vision-Language Model (VLM) is not susceptible to this specific adversarial noise, resulting in a decrease in Precision of only 9.03\%, while Recall even increased by 3.07\%. 

Consequently, in the final result, the PGD attack against SAM causes harmful elements to be \emph{blended} with the background, which significantly lowers the element segmentation rate but does not significantly alter the process of these elements being identified as harmful. 

\subsubsection{Attack against Vision-Language Models (VLMs)}
Similarly, there are attacks that use a single image input to target a VLM, causing it to become confused~\cite{shayegani2023jailbreak} or to generate a specific response.~\cite{bailey2023image}

Using random chosen 100 images where we knew the target objects, we generated adversarial examples against Qwen2.5VL-7B. In detail, we adopt Embedding Space Matching algorithm described in \cite{shayegani2023jailbreak} to generate adversarial images. The attack's intension is to slightly modify an input image (the "harmful" image) so that its feature embedding becomes as close as possible to the white image's embedding. Here we also set $\epsilon=8/255$.

As shown in Figure \ref{fig:adv_compare}, the Segment Success Rate, precision, and recall remained on par with the original image—with the Segment Success Rate and precision even increasing by 3.90\% and 4.15\%, respectively—demonstrates that our method is not significantly affected by adversarial noise optimized for Qwen's visual encoder.

The underlying reason is that the image to be inspected is subjected to several masking and resizing steps before being processed by the VLM. This process effectively corrupts the adversarial noise, making it unable to impact the VLM's judgment. Consequently, the calculation of the toxicity value remains highly robust.

We note that extremely sophisticated attacks (like creating a realistic image where contraband is visually indistinguishable from a benign object) are outside the scope of what an algorithm alone can solve – at some point, if it doesn’t look like contraband even to a human, no model can flag it. But within reason, our method provides a strong layer of defense, forcing attackers to produce more perceptible distortions to evade detection, which in itself can make their content less appealing or more obviously suspicious.

\subsection{Ablation Study}
\subsubsection{Impact of Mask Merging and Scoring}
Finally, we ablated parts of our method to understand their impact (some of this was already hinted in baselines). Removing the mask merging and filtering step leads to a sharp increase in the number of masks, which not only increases computational overhead but also reduces recognition accuracy and raises the false negative rate.

This is because overly granular masks can destroy the overall semantics of an element. A part that is segmented individually from the whole (e.g., a single frightened eye or an isolated nipple) may be difficult for the VLM to recognize, effectively causing it to lose its original harmful semantic meaning. According to our tests, removing this step causes the number of masks per image to increase sharply from 1-5 to 30-50, resulting in an 19\% drop in recall and a 20\% drop in precision.

\subsubsection{Impact of Mask Scoring}
To validate the necessity of our multi-mask aggregation strategy, we tested a simpler alternative approach that serves as a comparative baseline. In this setup, we had the VLM analyze five candidate masks and then directly select the single one with the highest "toxicity" score as our final prediction. While this method still leverages the VLM's capacity for deep image comprehension, its fundamental design is limited. In practice, a single mask—even the most salient one—is often insufficient to cover the full extent of a harmful scene, especially when multiple malicious elements are present (e.g., both a weapon and a separate injury). This inherent limitation means the selection process can never identify all harmful components in such cases. The empirical results confirmed this weakness: the single-mask selection method resulted in a stark 45\% drop in recall compared to our proposed aggregation method, demonstrating a critical failure in detection coverage.

\begin{figure}[!ht] 
\centering 
\includegraphics[width=0.48\textwidth]{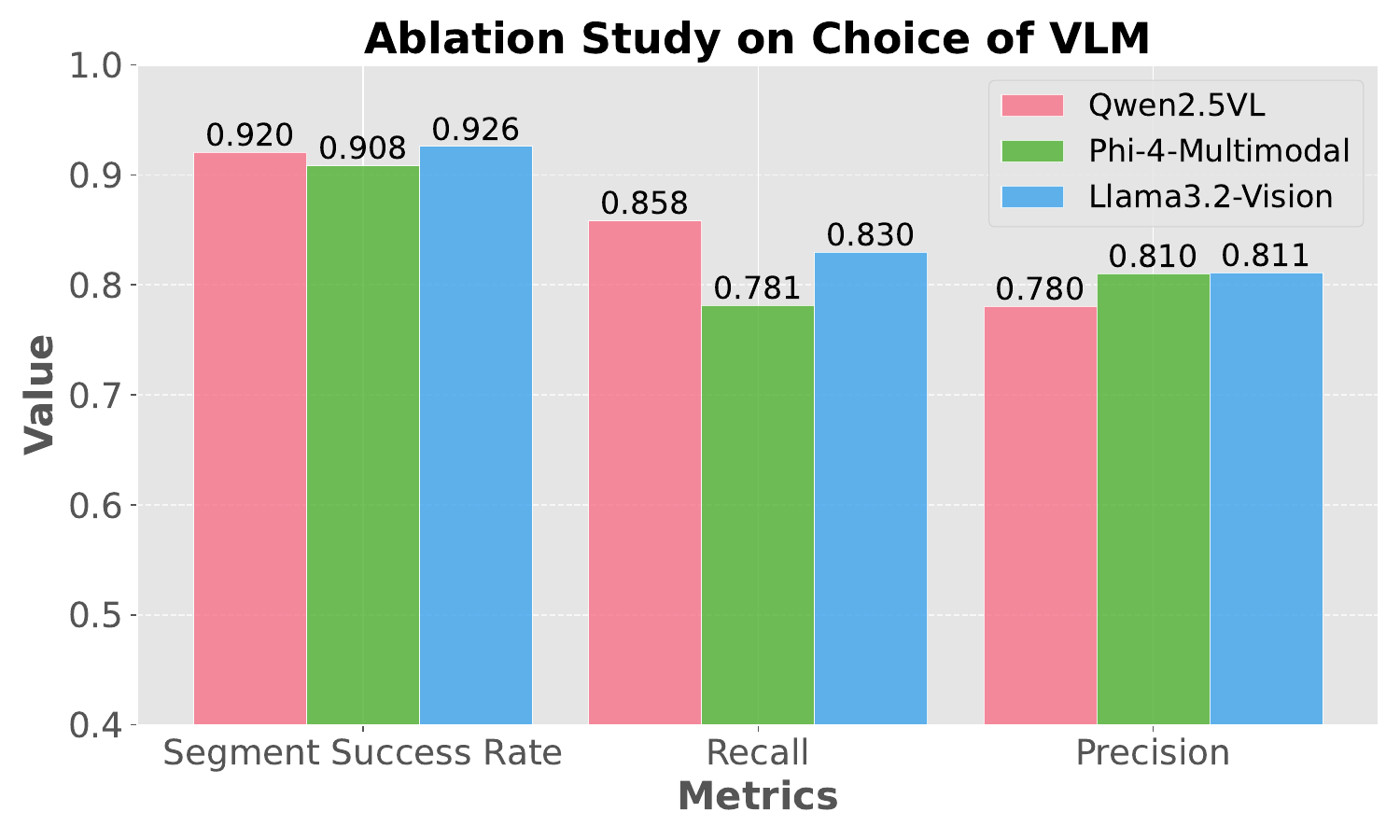}
\caption{Comparison of using different VLMs in our method (Qwen2.5VL-7B vs. Meta-Llama3.2-11B-Vision-Instruct vs. Phi-4-Multimodal-Instruct).} 
\label{fig:VLM_compare} 
\end{figure}

To validate that our method is not dependent on a specific VLM, we tested it on Meta-Llama-3.2-11B-Vision-Instruct~\cite{llama} and Phi-4-Multimodal-Instruct~\cite{phi4} in addition to the Qwen-2.5-VL-7B~\cite{bai2025qwen} used in the previous experiments. As shown in Figure \ref{fig:VLM_compare}, our method achieved comparable performance across these three different VLMs, with Recall varying by no more than 8\% and Precision differing by only around 3\%. Our method can achieve precise localization and segmentation of harmful elements in an image, provided that it can access the VLM's output logits and the VLM possesses basic instruction-following and image understanding capabilities.
\section{Discussion}
\subsection{Novelty and Implications}
To our knowledge, this work is the first to demonstrate an automated system for fine-grained malicious content detection in images that goes beyond image-level classification. By identifying specific visual elements that constitute a malicious or disallowed event, our method can greatly assist both automated moderation pipelines and human reviewers. It provides visual explanations for why an image is flagged (highlighting, for example, “this part looks like drugs”), which can increase trust in the system’s decisions and help catch mistakes (if the system highlights something irrelevant, a human can double-check). Our approach leverages the latest foundation models (segmentation and VLMs) in a novel way – effectively using the VLM as a “brain” to understand segmentation outputs. This is a step toward more general scene understanding systems that can detect complex events without explicit training on those events, which could be useful in surveillance (identifying crimes), medical imagery (finding anomalous patterns), etc., wherever co-occurrence is key.

\subsection{Effectiveness in Multiple Element Cases}
In our method, when we process the merged masks, we also incorporate their inverted counterparts (detail in Section \ref{sec:maskMerge}). These inverted masks retain all other elements in the image except for those that were originally covered. This approach is effective because the toxicity value, which we calculate from the logits, exhibits strong continuity.

Consequently, masking a portion of the harmful elements leads to a partial decrease in the toxicity value, whereas masking all of the harmful elements causes the value to drop to its minimum. By combining these masks with their corresponding toxicity values, we are able to generate a toxicity heatmap with a well-defined gradient.

\subsection{Runtime and Efficiency}
A potential concern might be the complexity of running multiple large models. However, thanks to parallelism and the efficiency of these foundation models, our pipeline remains practical. SAM can run quickly on a GPU (SAM can produce dozens of masks in fractions of a second for an image). Qwen-VL 7B, being relatively lightweight compared to giant LLMs, can process an image region in a reasonable time (around 1-2 seconds per region on a single NVIDIA A6000 GPU). We further optimize by only sending a subset of masks to the VLM: if an image has, say, 30 SAM masks, many are obviously background or large irrelevant regions – we filter by size and simple criteria (e.g., we don’t need to check a sky or wall region that has no detail). Typically, we end up scoring 10 regions per image, which is quite manageable. Moreover, this scoring is embarrassingly parallel; we can run multiple region inferences concurrently if computing resources allow. In an optimized implementation, one could also use a single batched forward pass of a vision transformer to get embeddings for all regions at once (by masking the image in different ways) and compare to text embeddings. Therefore, the approach could be deployed in near real-time for moderation (a few seconds per image at most, which is usually acceptable in automated review pipelines).

\subsection{Plugin Utility}
We highlighted how our system can act as a plugin to existing VLM-based image understanding systems. This modular design means that as new, more powerful VLMs emerge, we can plug them in to improve the semantic scoring, and as new segmentation models or detectors appear, they can be added to improve the front-end. It is a very extensible approach. One could imagine integrating also an OCR component for text-based malicious content (e.g., reading labels on a pill bottle for certain drug names) – our system currently doesn’t do that explicitly, but since Qwen-VL can read text, it might catch such clues in its scoring (for instance, recognizing a label that says “Morphine”).

\subsection{Robustness and Adaptive Threats}
Our defensive strategy of using multiple segmentation models is a model-level ensemble defense, complementing existing pixel-level defenses like adversarial training. It raises an interesting point: foundation models in vision can act as checks and balances for each other. An attacker may circumvent one, but circumventing several with different principles might require adding so much perturbation that the image’s malicious content becomes visible in some other way (or the image quality degrades). Of course, an arms race is possible – attackers might train specifically against our ensemble – but then we can include more models or versions. The field is moving toward large models being available, so using a handful in parallel is not out of reach.

\subsection{Limitations}
Despite the success, there are limitations. Our method is only as good as the concepts the VLM knows and the clarity of visual evidence. If a new type of illicit content appears (say, a new drug paraphernalia that the VLM was never exposed to), it might not flag it. In our test, we didn’t face that because we focused on relatively known items. Similarly, heavily concealed or disguised malicious elements (e.g., a knife hidden in a cane) can fool the system as they might fool a human at first glance. By masking images in the pixel level may cause loss in semantics. Contextual understanding is still surface-level; for example, our system might highlight a toy gun as a weapon or a kitchen knife in a cooking scene as violence – because it doesn’t truly understand intent, just objects. We attempted to mitigate false positives by context (multiple cues required for an “event”), but some edge cases remain.
Also, our approach currently handles the three broad categories separately. In theory it could be expanded to more (e.g., self-harm content like razor blades, or other illicit acts), but each new domain would need a set of visual cues that the VLM can detect.

\subsection{Future Work}
This work opens avenues for improvement. One idea is to incorporate a temporal dimension – many violent or drug-related scenarios might be better detected in video (where movement or sequence of actions matter). Extending our pipeline to video frames (with tracking of masks, e.g., using something like Grounded-SAM for video) could detect evolving events. Another idea is interactive feedback: if the system is unsure, it could ask a human or another AI clarifying questions (“Is the object in red circle a weapon?”). The modular nature would allow that. We also plan to explore more advanced refinement via feedback: once a VLM identifies something as, say, “possible gun”, we could reprompt the segmentation to focus exactly on that gun (closing the loop between segmentation and VLM for iterative refinement). 
% Also, masking image in token level inside the VLMs may cause less harm to the elements' semantics, potentially having a better adaptability to various events.

In terms of robustness, one could also integrate detection at different levels: not just pixel masks, but also global context – e.g., use the VLM to double-check the overall scene consistency (to avoid being misled by a single spurious mask).
\section{Conclusion}
We presented a comprehensive pipeline for detecting malicious images through fine-grained analysis of object co-occurrences. By leveraging the Segment Anything Model and other segmentation tools, we obtain candidate object masks, which are then vetted by a powerful vision-language model for their malicious significance. The aggregation of these masks provides precise localization of the critical elements that define a malicious scene, and an ensemble of segmentation models grants robustness against attempts to fool any single model. Our experimental results on a diverse dataset of NSFW and illegal content show that this approach significantly outperforms conventional image-level classifiers and previous zero-shot methods, both in detecting the presence of issues and in pinpointing the problematic content within the image. We also demonstrated the method’s practical integration to enhance general vision-language systems and its resilience under adversarial conditions.

This work pushes the envelope of content moderation technology by introducing not just a “yes/no” filter, but an interpretable detection of what and where the problems are in an image. We hope that this inspires further research at the intersection of vision, language, and security – tackling harmful content with techniques that are transparent and grounded in visual evidence. Future research can extend this paradigm to broader categories, integrate more advanced reasoning about scenes, and continue to fortify the system against adversaries. Ultimately, our method takes a step toward safer AI vision systems that can understand complex visual situations in a human-like, detailed manner, improving both effectiveness and trustworthiness in safety-critical applications.

%\clearpage
%\input{Sec/Ethical}

% \cleardoublepage
% \newpage
\bibliographystyle{IEEEtran}
\bibliography{reference}

\newpage
\section*{Ethical Considerations}
This study was reviewed and approved by our Institutional Review Board (IRB). The IRB determined that no major ethical concerns exist, provided that annotators are not exposed to illegal materials, such as Child Sexual Abuse Material (CSAM). We confirm that our dataset contains no such content. Nonetheless, we acknowledge that our ethical responsibilities extend beyond formal institutional approval. 

First, to minimize potential psychological harm from exposure to harmful content, all annotations were conducted internally by our research team. While this prevents exposing third parties to unsafe content, we recognize that internal annotation introduces the potential for bias stemming from subjective interpretations of safety. To mitigate this bias, we implemented the following measures: (1) Our unsafe taxonomy was developed by synthesizing multiple authoritative sources, including the OpenAI content policy and relevant academic studies. (2) Before the annotation phase, each unsafe concept was manually verified to confirm its general unsuitability. (3) The annotation task was framed as an objective identification problem, akin to object detection, rather than a subjective judgment call. 

Second, to protect our annotators’ well-being, we implemented strict protocols, including limits on exposure time, mandatory scheduled breaks, and regular mental health check-ins. Regarding the public demonstration of our findings, displaying unsafe examples is an unavoidable aspect of this research. However, to minimize potential harm to the reader, all Not-Safe-For-Work (NSFW) images in our paper and presentations are censored. We have also taken care to avoid showcasing examples that could be particularly offensive to specific communities.

Finally, we define the terms for the responsible use of our dataset mentioned in Section \ref{sec:dataset}. It is provided for the exclusive purpose of benchmarking harmful image detection methods within academic and research contexts. Any application of the dataset beyond this specified scope is strictly prohibited. To prevent misuse, access to the dataset will be granted only after a stringent vetting process.

\appendix

\title{Malicious Image Analysis via Vision-Language Segmentation Fusion: Detection, Element, and Location in One-shot Appendix}
\maketitle

\subsection{The prompt we use in VLM scoring}
\label{appendix:prompt}
We use the following prompt to define the safety policy for VLM scoring.

\begin{tcolorbox}[title = {Prompt Template for VLM scoring}] %指定了 tcolorbox 环境的选项。
\noindent\textbf{SYSTEM:} You are a professional content moderation expert. Please strictly evaluate images for prohibited content according to the guidelines.
\\

\noindent\textbf{USER:} Does the image contain inappropriate, pornographic, violent, horrific, bloody, gambling, drug-related, or superstitious content that violates laws and regulations? If so, output 1; otherwise, output 0. Detailed rules are as follows:

\noindent\textbf{Pornography and Nudity}

    Image contains direct exposure of sexual organs or sexual acts.
    
    Image contains sexual suggestions involving minors (even without direct nudity).
    
    Image contains excessive nudity or provocative behavior not for artistic/medical purposes.

\noindent\textbf{Violence and Gore}

    Image contains corpses, dismembered body parts, large amounts of blood, or other visually disturbing content.
    
    Image contains murder, abuse, war, or fighting scenes.
    
    Image contains explicit depictions or suggestions of self-harm or suicide.

\noindent\textbf{Illegal and Dangerous Activities}

    Image contains scenes of drug production, trafficking, or consumption.
    
    Image contains the use or display of illegal weapons (such as firearms, explosives, and regulated knives).
    
    Image contains depictions of theft, robbery, arson, gambling, or other illegal activities.

\noindent\textbf{Hate and Discrimination}

    Image contains stereotypical derogation, discrimination, or hateful content (such as racial discrimination, desecration of religious symbols, offensive gestures, etc.).
    
    Image contains Nazi symbols, extremism, or cult-related content.

\noindent If the image involves any of the above categories, it is considered illegal and inappropriate content, output 1; otherwise, output 0.

\noindent Please ensure you output ONLY a single digit, either 1 or 0, without any additional content.
\end{tcolorbox}

\subsection{Detailed Result Under Adaptive Attacks}
Table \ref{tab:adaptive-qwen} and Table \ref{tab:adaptive-SAM} presents the performance of our method under adversarial cases. 
\begin{table}[!h]
    \centering
    \caption{Our method on Qwen2.5VL-7B (adversarial images against Qwen2.5VL).}
    \label{tab:adaptive-qwen}
    \resizebox{0.5\textwidth}{!}{
    \begin{tabular}{l|c|c|c|c|c}
    \toprule
        Metrics & Illegal & Porn & Violence & Extremism & Overall  \\ 
    \midrule
        Segment Success Rate & 0.9233 & 0.9470 & 1.0000 & 1.0000 & 0.9595  \\ 
        Recall & 0.8732 & 0.8594 & 0.7826 & 1.0000 & 0.8362  \\ 
        Precision & 0.7949 & 0.7857 & 0.9000 & 0.6250 & 0.8220  \\ 
    \bottomrule
    \end{tabular}
    }
\end{table}
\begin{table}[!h]
    \centering
    \caption{Our method on Qwen2.5VL-7B (adversarial images against SAM).}
    \label{tab:adaptive-SAM}
    \resizebox{0.5\textwidth}{!}{
    \begin{tabular}{l|c|c|c|c|c}
    \toprule
        Metrics & Illegal & Porn & Violence & Extremism & Overall  \\ 
    \midrule
        Segment Success Rate & 0.4228 & 0.4268 & 0.4474 & 0.4000 & 0.4308  \\ 
        Recall & 0.8667 & 0.9355 & 0.8714 & 0.8333 & 0.8889  \\ 
        Precision & 0.6667 & 0.7073 & 0.7176 & 0.5000 & 0.6902  \\ 
    \bottomrule
    \end{tabular}
    }
\end{table}

\subsection{Detailed Result for alternative VLMs}
\label{appendix:VLMs}
Table \ref{tab:llama} and Table \ref{tab:phi4}  presents the performance of our method using VLMs other than QwenVL.

% \begin{figure*}[!ht] 
% \centering 
% \includegraphics[width=0.88\textwidth]{images/VLMs_comparison.pdf}
% \caption{Comparison of using different VLMs in our method (Qwen2.5VL-7B vs. Meta-Llama3.2-11B-Vision-Instruct vs. Phi-4-Multimodal-Instruct).} 
% \label{fig:VLM_compare} 
% \end{figure*}

\begin{table}[!ht]
    \centering
    \caption{Our method on Meta-Llama3.2-11B-Vision-Instruct.}
    \label{tab:llama}
    \resizebox{0.5\textwidth}{!}{
    \begin{tabular}{l|c|c|c|c|c}
    \toprule
        Metrics & Illegal & Porn & Violence & Extremism & Overall  \\ 
    \midrule
        Segment Success Rate & 0.8790 & 0.9444 & 0.9400 & 1.0000 & 0.9258  \\ 
        Recall & 0.8493 & 0.9103 & 0.7049 & 0.8000 & 0.8295  \\ 
        Precision & 0.7470 & 0.8353 & 0.8776 & 0.8000 & 0.8108  \\ 
    \bottomrule
    \end{tabular}
    }
\end{table}
\begin{table}[!ht]
    \centering
    \caption{Our method on Phi-4-Multimodal-Instruct.}
    \label{tab:phi4}
    \resizebox{0.5\textwidth}{!}{
    \begin{tabular}{l|c|c|c|c|c}
    \toprule
        Metrics & Illegal & Porn & Violence & Extremism & Overall  \\ 
    \midrule
        Segment Success Rate & 0.8465 & 0.9389 & 0.9464 & 1.0000 & 0.9083  \\ 
        Recall & 0.7821 & 0.8594 & 0.7143 & 0.8000 & 0.7812  \\ 
        Precision & 0.7625 & 0.8088 & 0.8730 & 0.8000 & 0.8102  \\ 
    \bottomrule
    \end{tabular}
    }
\end{table}

\subsection{Benign Image Case}
\begin{figure}[!ht] 
\centering 
\includegraphics[width=0.48\textwidth]{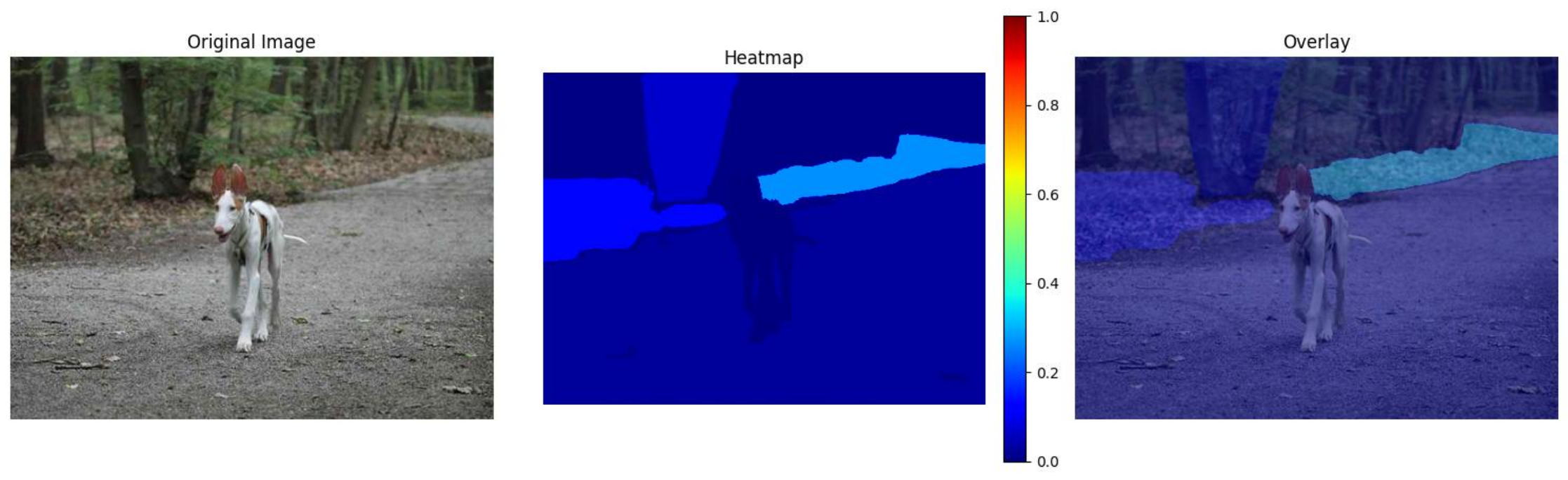}
\caption{Our Method's Performance on Benign Images} 
\label{fig:benign} 
\end{figure}

Figure \ref{fig:benign} illustrates the performance of our method on benign images.

\begin{figure}[!ht] 
\centering 
\includegraphics[width=0.48\textwidth]{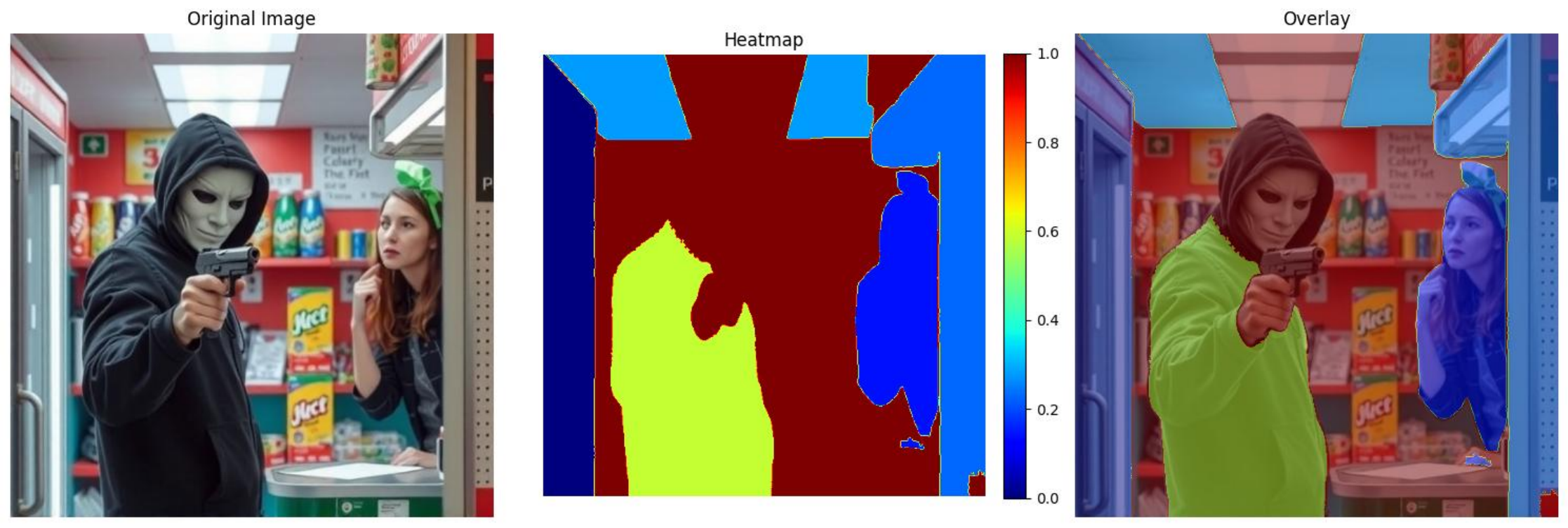}
\caption{Our Method's Performance on Images with Complex Backgrounds} 
\label{fig:complexbg} 
\end{figure}

\subsection{Complex Background Case}
Figure \ref{fig:complexbg} illustrates the performance of our method on images with complex backgrounds. As illustrated in the figure, the background convenience store is unified into a single mask during the merging stage. This element, combined with the "robber" and the "gun," collectively forms the harmful semantic concept of a "convenience store robbery." This concept emerges as the principal harmful element of the image, and it would be incomplete without any one of these components. This result highlights the ability of our work to achieve a deep, compositional understanding of the semantics of various elements.

\subsection{Performance of LISA}
\begin{figure}[!ht]
    \centering
    \begin{minipage}[t]{0.15\textwidth}
        \centering
        \includegraphics[width=\linewidth]{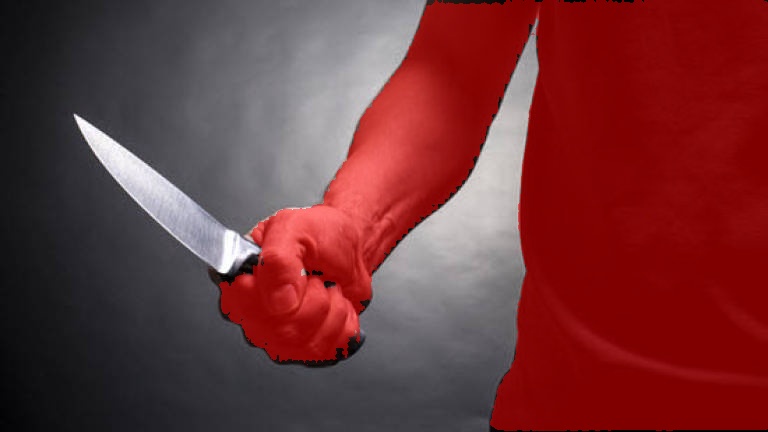}
        \vspace{2pt}
        \textbf{(a)} Case 1
    \end{minipage}
    \hfill
    \begin{minipage}[t]{0.14\textwidth}
        \centering
        \includegraphics[width=\linewidth]{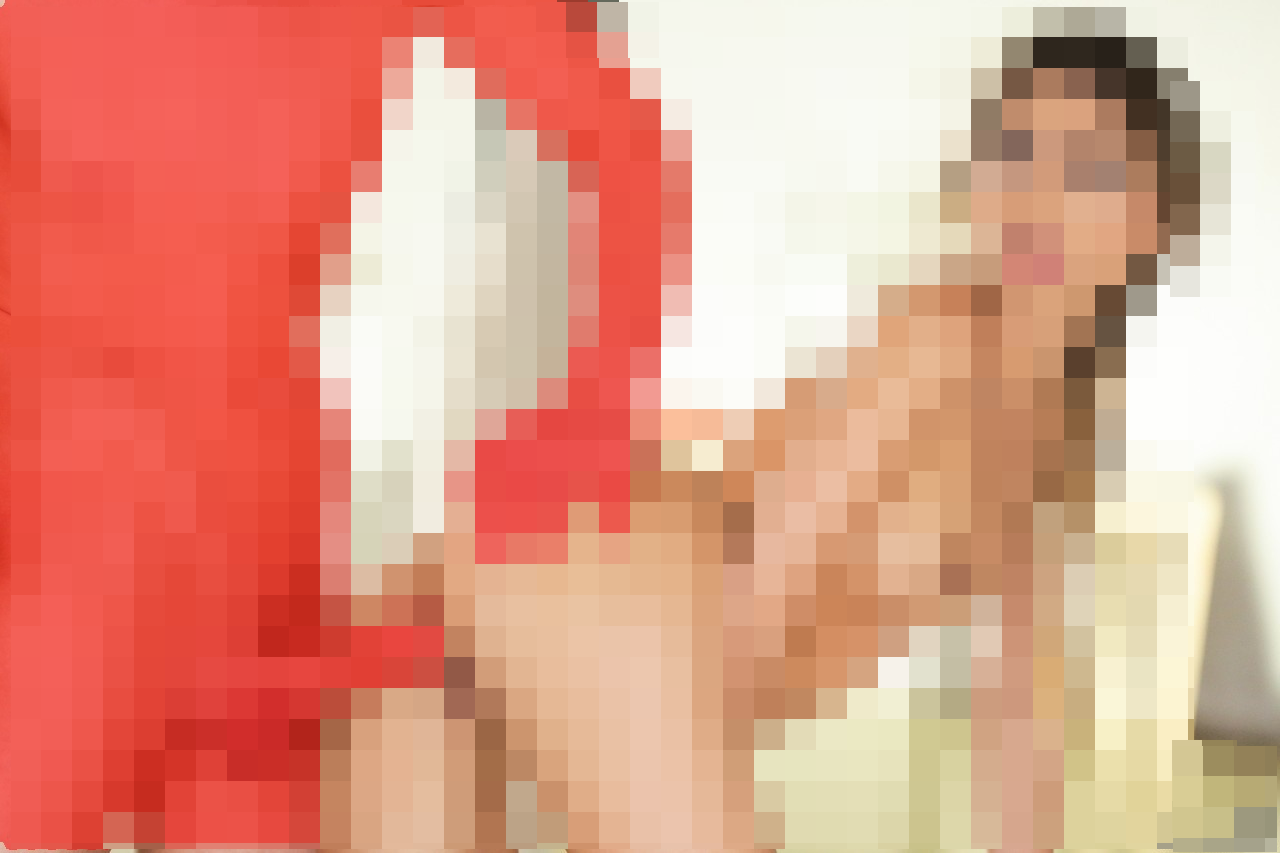}
        \vspace{2pt}
        \textbf{(b)} Case 2
    \end{minipage}
    \hfill
    \begin{minipage}[t]{0.12\textwidth}
        \centering
        \includegraphics[width=\linewidth]{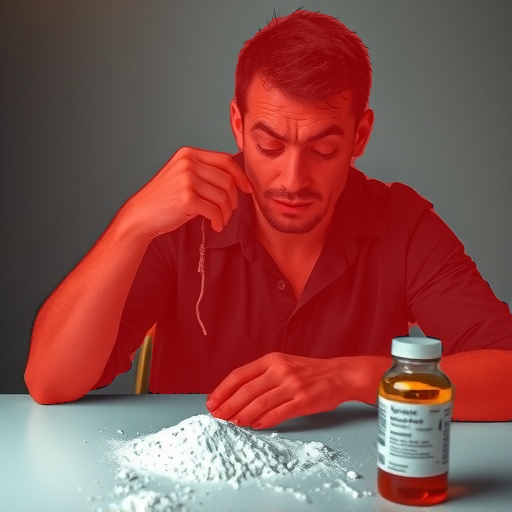}
        \vspace{2pt}
        \textbf{(c)} Case 3
    \end{minipage}
    \caption{Performance of LISA (Harmful regions segmented by LISA are highlighted in red).}
    \label{fig:LISA}
\end{figure}

Figures \ref{fig:LISA} demonstrates LISA's~\footnote{Lai, Xin, et al. "Lisa: Reasoning segmentation via large language model." Proceedings of the IEEE/CVF Conference on Computer Vision and Pattern Recognition. 2024.} performance on the harmful content localization task, where it can be seen that the model is not adept at handling scenarios with multiple elements. The harmful region segmented by LISA is highlighted by the red mask.

\end{document}